\documentclass[10pt,peerreview,draftclsnofoot,onecolumn,a4paper]{IEEEtran}
\usepackage{amsmath,amssymb}
\usepackage{microtype,charter}
\usepackage[linesnumbered,boxed]{algorithm2e}
\usepackage{graphicx,epsfig,url,epstopdf}
\usepackage[caption=false,font=footnotesize]{subfig}

\def\F{\mathbf{F}}
\def\fr{f_\text{r}}
\def\fc{f_\text{c}}
\def\frc{f_\text{rc}}
\def\fcr{f_\text{cr}}
\def\boldtheta{\boldsymbol{\theta}}
\newtheorem{theorem}{Theorem}[section]
\newtheorem{proposition}[theorem]{Proposition}
\newtheorem{corollary}[theorem]{Corollary}
\newcommand{\vertiii}[1]{{\left\vert\kern-0.25ex\left\vert\kern-0.25ex\left\vert #1
\right\vert\kern-0.25ex\right\vert\kern-0.25ex\right\vert}}

\title{Fast Separable Non-Local Means}
\author{Sanjay Ghosh and Kunal N. Chaudhury \thanks{The authors are with the Department of Electrical Engineering, Indian Institute of Science, Bangalore, India. Correspondence: kunal@ee.iisc.ernet.in. This work was supported in part by a Startup Grant from the Indian Institute of Science.}}

\begin{document}
\maketitle
\begin{abstract}
We propose a simple and fast algorithm called PatchLift for computing distances between patches (contiguous block of samples) extracted from a given one-dimensional signal. PatchLift is based on the observation that the patch distances can be efficiently computed from a matrix that is derived from the one-dimensional signal using lifting; importantly,  the number of operations required to compute the patch distances using this approach does not scale with the patch length. We next demonstrate how PatchLift can be used for patch-based denoising of images corrupted with Gaussian noise. In particular, we propose a separable formulation of the classical Non-Local Means (NLM) algorithm that can be implemented using PatchLift. 
We demonstrate that the PatchLift-based implementation of separable NLM is few orders faster than standard NLM, and is competitive with existing fast implementations of NLM.
Moreover, its denoising performance is shown to be  consistently superior to that of NLM and some of its variants, both in terms of PSNR/SSIM and visual quality.
\end{abstract}

\textbf{Keywords}: Non-local means, denoising, patch distance, fast algorithm, separable filtering, lifting.

\section{Introduction}
\label{introduction}

The Non-Local Means (NLM) algorithm was introduced by Buades, Coll, and Morel \cite{Buades2005} for denoising natural images corrupted with additive Gaussian noise. Two key innovations of NLM are the effective use of non-local correlations in natural images, and the use of patches (instead of single pixels) to robustly measure photometric similarity.
The conceptual simplicity of NLM, coupled with its excellent denoising performance,  triggered a series of work on the use of patch-based models for image denoising \cite{Kervrann2006,KSVD,BM3D,Giloba2008,Chatterjee2012}. Some of these methods currently provide state-of-the-art results 
for a wide class of natural images. We refer the readers to \cite{Chatterjee2012,BCM2010} for an exhaustive account of such patch-based algorithms. While NLM is no longer the top algorithm for image denoising, it nevertheless continues to be of interest due to its simplicity, decent denoising performance, and the availability of several fast implementations \cite{Wang2006}-\cite{Dauwe2008}.

The non-local means of an image $f = \{f(i) : i \in \Omega\}$, where $\Omega  = \big\{ i=(i_1, i_2):  1\leq i_1,i_2 \leq N\}$, is given by \cite{Buades2005}
\begin{equation}
\label{NLM}
\text{NLM}[f](i)  = \frac{\sum_{j \in S(i)} w_{ij} f(j)}{\sum_{j \in S(i)} w_{ij} } \qquad (i \in \Omega).
\end{equation}
To exploit long-distance correlations in natural images, the original proposal in [1] was to set the neighborhood $S(i)$ to be the whole image. In practice, one restricts $S(i)$ to a sufficiently large window of size $(2S + 1) \times (2S + 1)$ centred at $i$. 
The weights $\{w_{ij} : i \in \Omega, j \in S(i) \}$ are given by 
\begin{equation}
\label{weights}
 w_{ij} = \exp \Big( - \frac{1}{h^2} \sum_{k \in P} G_{\alpha}(k) \big( f(i+k) - f(j+k) \big)^2 \Big).
\end{equation}
Here, $h$ is a smoothing parameter, $P= [- K, K]^2$ is a patch of size $(2K + 1) \times (2K + 1)$ centered at the origin, and
$G_{\alpha}$ is a two-dimensional Gaussian kernel:
\begin{equation*}
G_{\alpha} (k_1,k_2) =  \exp\Big( - \frac{1}{2 \alpha^2} (k_1^2+k_2^2)\Big).
\end{equation*}
We  note that in several recent papers on non-local means, such as \cite{Darbon2008,Tasdizen2008,Tasdizen2009} and \cite{WTNN2013}, the authors have used a box kernel instead of a Gaussian kernel. That is, $G_{\alpha}(k)$ are set to unity for $k \in P$ in \eqref{weights} in this case. We will consider both Gaussian and box kernels in this paper. Along with the non-local averaging in \eqref{NLM}, it is the use of patches that makes NLM more robust than pixel-based neighborhood filters \cite{Yaroslavsky1985,Smith1997,Tomasi1998}.

\subsection{Fast Non-Local Means}

The direct implementation of NLM is computationally demanding and slow in practice \cite{Buades2005}. In particular, the  computation of \eqref{NLM} requires $O(N^2S^2K^2)$ operations for an  $N \times N$ image. Several computational tricks and tradeoffs have been proposed to speedup NLM. These can broadly be classified into the following classes:

$\bullet$ Fast Weight Computation: In this class of algorithms, the  patch distances are computed rapidly using convolutions and FFTs, possibly at the cost of added storage. 
For instance, a fast method using integral images \cite{VJ2001} and FFT computations was proposed in \cite{Wang2006}. This method is tailored to work with the box profile and cannot be used for the Gaussian kernel. The authors, however, remark in the paper that the PSNR obtained using the box profile is generally less than that obtained using the Gaussian profile. The authors in \cite{KUD2009} use integral images along with a multi-resolution image pyramid to speed up the weight computation. A fast algorithm for computing the patch weights in constant time (independent of the patch size) was proposed in \cite{Darbon2008}. More recently, an exact and simple implementation of NLM using convolutions was described in \cite{Condat2010}. A probabilistic weighting scheme was proposed in \cite{WTNN2013} which is faster than original NLM and provides higher PSNRs.

$\bullet$ Neighborhood Selection: A leading factor in the computational load of NLM is the size of the search window. Naturally, a means of speeding up (and possibly improving) NLM is to use  image priors to perform neighbourhood selection, that is, to preselect similar neighborhoods based on some criteria. For instance, the authors in \cite{Sapiro2005} used  local gray values and gradients to eliminate  neighborhoods with small correlations. A parallelized blockwise implementation was proposed in \cite{Coupe2008}, which includes a technique to automatically tune the smoothing parameter $h$ based on the mean and variance of the pixel intensities. In a different direction, a fast and accurate means of preselecting similar patches by arranging the noisy image in a cluster tree was proposed in \cite{Brox2008}. This method was shown to perform well on images with regular, textured patterns.  The statistical nature of the additive noise has also been explored to improve neighborhood selection. For example, the first three statistical moments of the noisy image have been used in \cite{Dauwe2008} to derive a threshold for rejecting dissimilar patches which leads to a reduction in the overall runtime. A related approach was proposed in \cite{earlytermination2010}, where dissimilar patches are dropped if the distortion between patches exceed an expected threshold, which is calculated from the additive noise strength. 

$\bullet$ Dimensionality Reduction: In a different direction, it was shown in  \cite{Tasdizen2008,Tasdizen2009,Orchard2008} that the run time and the denoising performance of NLM can be improved by first projecting the patches into a lower dimensional subspace and then computing the patches distances in this subspace. Singular value decomposition and principal component analysis were respectively used in these papers to reduce the dimension. 

$\bullet$ Efficient Data Structure: In \cite{Adams2009,Adams2010}, the NLM computation was posed as a multi-dimensional filtering and efficient data-structures were used to speedup the computation. In particular, a kd-tree-based sampling for accelerating a broad class of non-linear filters was proposed in \cite{Adams2009} that included NLM as a special case. Later, another high dimensional data structure was presented in \cite{Adams2010} that resulted in the fastest known implementation of non-local means for dimensionality ranging between 5 and 20.

In addition to the above mentioned work, researchers have proposed to replace the usual square patches in classical NLM with various shapes (disk, band, etc.) so as to exploit the local geometry of the image. For example, a fast algorithm was proposed in \cite{Deledalle2012} which can work with such patch shapes. This algorithm also used FFT-based computations to improve the runtime.
Recently a learning-based NLM variant was proposed in \cite{BC2014}, where patch dictionaries are used to speed up the distance computation.

\begin{table}
\caption{Complexity of different algorithms for NLM and its variants. $S$  and $K$ denote the half-width of the search window and the half-width of the patch respectively, and $N \times N$ is the image size. Also indicated in the table is whether a particular algorithm works with both box and Gaussian kernels.}
%The computations were performed using Matlab on a 3.40 GHz Intel quad-core machine with 16 GB memory. }
\vspace{2mm}
\centering
\begin{tabular}{| c | c | c | c| c |}
\hline
Method          & Complexity   & Gaussian  & Box  %& Addl. Comments
\\
\hline
Buades \cite{Buades2005}       & $O( N^2 S^2 K^2 )$      & \checkmark    &  \checkmark       \\
Vignesh \cite{earlytermination2010} & $O( N^2 S^2 K^2 )$      & \checkmark    &  \checkmark    \\
 Mahmoudi \cite{Sapiro2005}    & $O( N^2 S^2 K^2 )$      & \checkmark    &  \checkmark    \\
Brox \cite{Brox2008}        & $O(N^2K^2 \text{log} N)$ & \checkmark    &  \checkmark \\
 Wang \cite{Wang2006}      &  $O( N^2 S^2 \text{log} S)$       & $\times$     & \checkmark          \\
Darbon \cite{Darbon2008}     & $O( N^2 S^2 )$     & $\times$     & \checkmark       \\
Condat \cite{Condat2010}     & $O( N^2 S^2 )$     & $\times$     & \checkmark          \\
Proposed             & $O( N^2S)$       & \checkmark     & \checkmark          \\
\hline
\end{tabular}
\label{complexity}
\end{table} 

\subsection{Present Contribution}

The present work is based on the idea of separable filtering which is common in the image processing literature. For example, separable formulations of bilateral and median  filtering have been reported in \cite{N1981,Pham2005}. A separable formulation of NLM was recently proposed in \cite{KLCLKK2011}. We note that the above non-linear filters are not separable as such; that is, one cannot perform the filtering by first processing the rows and then the columns (as can be done for linear filters with separable kernels). In fact, the result of row-filtering followed by column-filtering is generally different from that obtained by reversing the operations. In this work, we propose to take both these possibilities into account and express the final denoised image as an optimal combination of these primitives. The optimality is in the sense of a certain surrogate of the mean-squared error that is popularly referred to as Stein's unbiased risk estimator \cite{S1981,BL2007}. The resulting optimization problem reduces to solving a $2 \times 2$ linear system. The present approach for separable NLM is different from that in \cite{KLCLKK2011} and provides  much better denoising results.

The main novelty of the paper, however, is the proposal of an algorithm that can reduce the complexity of one-dimensional NLM from $O(NSK)$ to $O(NS)$, where $N$ is the length of the signal. This is based on the observation that the patch distances involved in the NLM of a one-dimensional signal can be computed from $O(NS)$ entries of a specially designed matrix. This matrix is obtained by applying a (box or Gaussian) filter along the subdiagonals of a matrix which is derived via \textit{lifting}, namely, through the tensor product of the signal with itself. As is well-known, box and Gaussian filtering can be performed using $O(1)$ operations with respect to the filter length (e.g., see \cite{D1992,YV1995}). The proposed algorithm as a result requires $O(NS)$ operations to compute the full set of patch distances in NLM. To the best of our knowledge, the observation that lifting can be used for efficiently computing patch distances is novel. We next use this lifting-based algorithm to develop a fast separable formulation of NLM for grayscale images (as described in the above paragraph). The complexity of the proposed separable formulation of NLM is $2N \times O(NS) = O(N^2S)$ for an $N\times N$ image, which is substantially smaller than the $O(N^2S^2K^2)$ complexity of standard NLM. In practice, the proposed method is at least $300$ times faster than original NLM when $S=20$ and $K=5$.  Moreover, as will be shown in the sequel, the PSNR obtained using separable NLM is consistently larger than that obtained using original NLM and some of its variants. A comparison of the complexity of various algorithms for NLM (and its variants) is provided in Table \ref{complexity} for reference.

\subsection{Organization}

The rest of the paper is organized as follows. We present the fast algorithm for one-dimensional NLM in Section \ref{sec:fast 1d}. The main component of this section is a constant-time algorithm for computing patch distances that has a particularly simple implementation. In Section \ref{sec: fast separable}, we introduce a separable formulation of the standard NLM algorithm. In particular, we demonstrate how the results of row-column and column-row filtering (which are generally different) can be combined in an optimal fashion (at a marginal overhead).
We implement this separable formulation of NLM using the fast algorithm from Section \ref{sec:fast 1d}. We provide several simulation results in Section \ref{sec:results} and compare the proposed method with existing fast implementations of NLM. We end the paper with some concluding remarks in Section \ref{sec:conclusion}.

\section{Fast One-Dimensional NLM}
\label{sec:fast 1d}

Consider a one-dimensional signal $f = \{f(1), \ldots,f(N)\}$ of length $N$ (corresponding to a row or a column of an image).
The one-dimensional counterpart of a patch is simply the collection of samples $\{f(i + k), k \in [-K,K]\}$, where $i$ is the centre of the patch and $2K+1$ is its length.
%We assume here and henceforth that the signal has been appropriately extended beyond its boundaries.
Consider two patches  centred at $i$ and $j$. We define the weighted Euclidean distance between these patches to be
\begin{equation}
\label{distanceSquared}
d_{ij}^2 =  \sum_{k=-K}^K g_{\beta}(k) \big( f(i + k) - f(j + k) \big)^2,
\end{equation}
where $g_{\beta}$ is a one-dimensional Gaussian kernel:
 \begin{equation*}
g_{\beta} (k) =  \exp\left( - \frac{k^2}{2 \beta^2} \right).
\end{equation*}
As in the two-dimensional setting, we also consider the possibility of $g_{\beta}$ being a box kernel. The one-dimensional analogue of \eqref{NLM} is given by
\begin{equation}
\label{1dNLM}
\hat{f}(i)  =\frac{\sum_{j = i-S}^{i+S} w_{ij} f(j)}{\sum_{j = i-S}^{i+S} w_{ij} },
\end{equation}
where $w_{ij} = \exp(- d_{ij}^2 /h^2)$. For any given $i$, note that a total of $2S+1$ distance computations are required. Since each distance computation involves $O(K)$ operations, the total cost of computing  the weights $\{w_{ij}:  1 \leq i \leq N, i-S \leq j \leq i+S\}$ in \eqref{1dNLM}  is $O(NSK)$.

\subsection{PatchLift}
\label{sec:patch dist}

We now explain how the redundancy involved in computing the patch distances (and hence the weights) can be exploited in a principled fashion. 
Notice that if we expand the terms in \eqref{distanceSquared}, we get a sum involving the product of signal samples. This makes it difficult to exploit the overlap between the adjacent patches. However, this non-linear dependence (on the samples) can be transformed into a linear one by using an appropriate data structure.
In particular, we consider the $N \times N$ matrix $\F$ obtained by taking tensor product of the signal with itself, namely,
\begin{equation}
\label{tensor}
\F(i,j) = f(i) f(j), \quad \quad  1 \leq i,j \leq N.
\end{equation}
We then smooth $\F$ by filtering its subdiagonals using $g_{\beta}$:
\begin{equation}
\label{convn}
\overline{\F}(i,j) =  \sum_{k=-K}^K g_{\beta}(k) \F(i + k,j + k).
\end{equation}
Using \eqref{tensor} and \eqref{convn}, we can then write \eqref{distanceSquared} as
\begin{equation}
\label{lift}
d_{ij}^2 = \overline{\F}(i,i) + \overline{\F}(j,j) -2 \overline{\F}(i,j).
\end{equation}
Notice that we have effectively transferred the non-linearity in \eqref{distanceSquared} onto $\overline{\F}$.
In particular, the patch distances can now be computed using just three samples of $\overline{\F}$, one multiplication, and two additions.

\begin{figure}
\centering
\includegraphics[width=0.5 \linewidth]{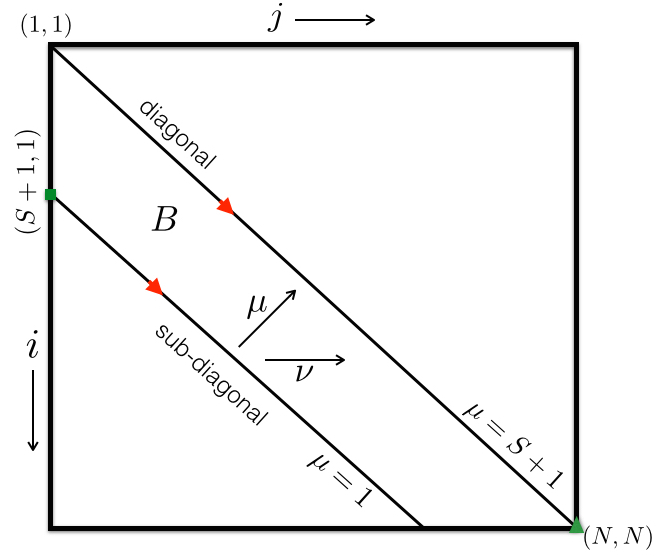}
\caption{The square shown above is the domain of $\F$, and $(i,j)$ are its indices. The one-dimensional sequence $V$ is formed by stacking the subdiagonals of $\F$ in order; $\mu$ is used to index the subdiagonals, while $\nu$ indexes the position within a given subdiagonal. The correspondence $(i,j) \leftrightarrow (\mu,\nu)$ is given by \eqref{correspondence}. The start and the end of $V$ are respectively marked with a green square and a triangle; the domain of $V$ is the band $B$ given by \eqref{band}. The red arrows indicate the direction along which the samples of $V$ are collected from $\F$.} 
\label{packing}
\end{figure}

Note that we only require a portion of $\overline{\F}$ around the diagonal for computing the distances using \eqref{lift}. Moreover, due to the symmetry in \eqref{tensor}, it suffices to consider the samples $\overline{\F}(i,j)$  for which $i \geq j$. In particular, we require the samples $\overline{\F}(i,j)$ where
$(i,j)$ takes values in the band
\begin{equation}
\label{band}
B = \Big\{(i,j):  \ 1 \leq i,j \leq N, \ i \geq j, \  |i-j| \leq S  \Big\}.
\end{equation}
It is not difficult to see from \eqref{convn} that we do not need the entire $\F$  for computing $\{\overline{\F}(i,j): (i,j) \in B\}$; we only require the samples $\{ \F(i,j): (i,j) \in B \}$. In other words, we need $O(NS)$ samples of $\overline{\F}$ for computing the patch distances, which in turn can be obtained from $O(NS)$ samples of $\F$ using \eqref{convn}. 

\subsection{Implementation}

We can implement PatchLift efficiently by packing the subdiagonals of $\F$ into a sequence and convolving the resulting sequence with $g_{\beta}$. A small detail in doing so is that we need to pack $K$ zeros between successive subdiagonals to avoid interference between the successive filtering (recall that the support of $g_{\beta}$ is $[-K,K]$). The samples of $\overline{\F}$ can then be read off directly from the filtered sequence. More specifically, suppose we denote the indices of the subdiagonals using $\mu$, where $\mu$ takes values from $1$ to $S+1$. In particular, $\mu=S+1$ corresponds to the diagonal, and $\mu=1$ corresponds to the subdiagonal that is farthest from the diagonal (see Figure \ref{packing}). We linearly index the elements within a given subdiagonal using $\nu$. For a subdiagonal with index $1 \leq \mu \leq S+1$, $\nu$ varies from $1$ to $N - S -1 +\mu$, that is, the length of the subdiagonal is $N - S -1 +\mu$. It can be verified that the correspondence between the indices $(i,j) \in B$ and the indices $(\mu,\nu)$  
is given by 
\begin{equation}
\label{correspondence}
j = \nu,\  i = S+1-\mu+\nu  \quad \Leftrightarrow \quad  \nu = j, \ \mu= S+1-i+j.
\end{equation}
We now form a  one-dimensional sequence $V$ by stacking the subdiagonals of $\F$ in order, and packing $K$ zeros between successive subdiagonals. The linear position of a point $(i,j) \in B$ with respect to $V$ is then given by
\begin{equation*}
l_{ij} =   (\mu - 1)K + \sum_{t = 1}^{\mu -1} (N-S-1+t)  + \nu,
\end{equation*}
where $\mu$ and $\nu$ are given by \eqref{correspondence}. 
The first term accounts for the zero-packings, the second term counts the number of elements in subdiagonals $1,\ldots,\mu-1$, and the third term gives the position within the $\mu$-th subdiagonal. After simplification and using \eqref{correspondence}, we get
\begin{equation}
\label{packing_1}
l_{ij} = \frac{1}{2}(S-i+j)(2N + 2K - S - 1 - i + j) + j.
\end{equation}
In other words, we have
\begin{equation*}
\F(i,j) = V(l_{ij}) =f(S+ 1 - \mu + \nu)f(\nu).
\end{equation*}
It is not difficult to see that we can now express \eqref{convn} using the one-dimensional convolution
 \begin{equation}
 %\label{con}
 \overline{V} = V \ast g_{\beta}.
\end{equation}
Notice that we can get the total length of $V$ simply by setting $i=N$ and $j=N$ in \eqref{packing_1}. In particular, the length of $V$ is $L=(S+1)(N-S/2) + KS$. 
The above process of setting up $V$ and computing $\overline{V}$ is summarized in Algorithm \ref{algo1}.
Notice that the zero-paddings of length $K$ are introduced in step.
%\ref{increment}.

\IncMargin{1mm}
\begin{algorithm}
\KwData{Signal $f(1),\ldots,f(N)$, kernel $g_{\beta}$, and $K,S$.}
\KwResult{Sequence $\overline{V}$ of length $(S+1)(N-S/2) + KS$.}
\textbf{Initialize}: Null sequence $V$ of length $L$, and $l=0$\;
\For{$\mu = 1, 2, \ldots, S+1$}{
     \For{$\nu = 1, 2, \ldots, N - S -1 +\mu$}{
       $l = l+1$\;
        $V(l) = f(S+1-\mu+\nu)  f(\nu)$\;
}$l = l + K$\; \label{increment}
}
$\overline{V} = V \ast g_{\beta}$. \label{con}
\caption{Computation of $\overline{V}$.}
\label{algo1}
\end{algorithm}
\DecMargin{1mm}

\IncMargin{2mm}
\begin{algorithm}
\KwData{Signal $f(1),\ldots,f(N)$, and parameters $K,S,h$.}
\KwResult{NLM of $f$ given by \eqref{1dNLM}.}
Compute  $\overline{V}$ from $f, K,$ and $S$ using Algorithm \ref{algo1}\;
\For{$i=1,\ldots,N$} {
$P = 0$\;
$Q = 0$\;
\For{$j =  i - S,\ldots,i+S$}{
Compute $d_{ij}^2$ using \eqref{dist}\;
$w = \exp(- d_{ij}^2/h^2)$\;
$P=P+w  f(j)$\;
$Q=Q+w$\;
}
$\hat{f}(i) =P/Q$\;
}
\caption{PatchLift-NLM.}
\label{algo2}
\end{algorithm}
\DecMargin{2mm}

It is clear that $V$ has $L = O(NS)$ elements. Therefore, if the convolution in step \ref{con} can be computed in $O(1)$ operations, then the complexity of setting up $V$ and computing $\overline{V}$ would be $O(NS)$. As is well-known, this is indeed the case when $g_{\beta}$ is a box or a Gaussian kernel. We have thus established the following fact.

\begin{proposition} 
\label{prop}
The distances $\{d_{ij}:  1 \leq i \leq N, i-S \leq j \leq i+S\}$ given by \eqref{distanceSquared} can be computed using $O(NS)$ operations for any arbitrary $K$ for box and Gaussian kernels.
\end{proposition}

In particular, we have the following recursion for the box kernel:
\begin{equation}
\label{recursion}
\overline{V}(l) = \overline{V}(l-1) + \left[V(l+K) -V(l-K -1) \right].
\end{equation}
Thus, given $ \overline{V}(l)$, we can compute $\overline{V}(l+1)$ using just two additions for any arbitrary $K$. The recursive implementation of the Gaussian filter is somewhat more involved and the details can be found in \cite{D1992,YV1995}. We will henceforth refer to the proposed approach of computing patch distances using lifting as PatchLift.

\subsection{PatchLift-NLM}

\begin{figure}
\centering
\includegraphics[width=0.7\linewidth]{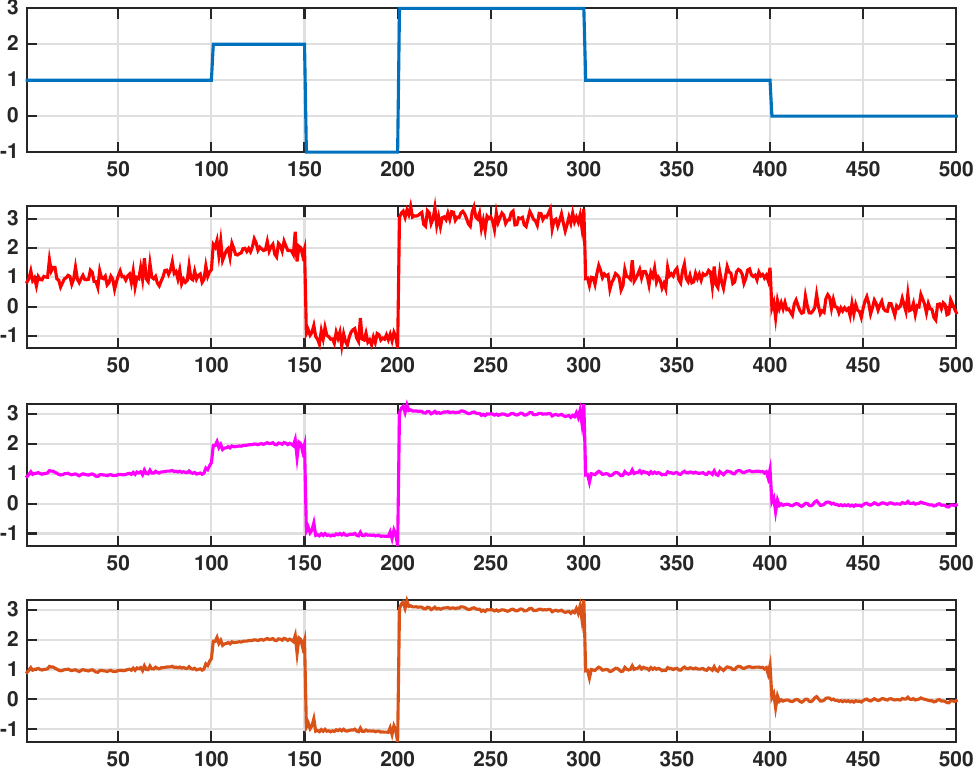}
\caption{Denoising of a one-dimensional signal using NLM and PatchLift-NLM. \textbf{Row 1}: Clean signal; \textbf{Row 2}: Clean signal corrupted with additive white Gaussian noise ($\sigma=0.2$); \textbf{Row 3}: Output of NLM computed using \eqref{1dNLM}; \textbf{Row 4}: Output of PatchLift-NLM computed using Algorithm \ref{algo2}. The parameters used are $S=10, K=5$, and $h=1$. PatchLift-NLM is about 30 times faster in this case, and the mean-squared-error between the output of NLM and PatchLift-NLM is of the order $10^{-17}$.} 
\label{scanLineDenoising}
\end{figure}

Having computed $\overline{V}$ using Algorithm \ref{algo1}, we can use it to determine the patch distances. Note that, by construction, $\overline{\F}(i,j) = \overline{V}(l_{ij})$ for $(i,j) \in B$. We can therefore write \eqref{lift} as
\begin{equation}
\label{dist}
d_{ij}^2 = 
\begin{cases}
     \overline{V}(l_{ii}) + \overline{V}(l_{jj}) -2 \overline{V}(l_{ij})   \hspace{0.5 cm} &  i \geq j , \\
    \overline{V}(l_{ii}) + \overline{V}(l_{jj}) -2 \overline{V}(l_{ji}) \hspace{0.5 cm} & j > i.
  \end{cases}
\end{equation}
By using the patch distances obtained using \eqref{dist}, we can compute \eqref{1dNLM} using another $O(NS)$ operations. We thus have the following result as a consequence of Proposition \eqref{prop}.
\begin{corollary}
\label{coro}
The NLM filtering in \eqref{1dNLM} can be computed using $O(NS)$ operations for any arbitrary $K$ for box and Gaussian kernels.
\end{corollary}
We will refer to the proposed implementation of NLM using PatchLift as PatchLift-NLM. This is summarized in Algorithm \ref{algo2}. 
For completeness, we present the results of denoising a synthetic signal using NLM and PatchLift-NLM in Figure \ref{scanLineDenoising}.
The two outputs are visually indistinguishable as expected, while a speedup by a factor of $30$ is obtained using PatchLift-NLM when $K=5$.

\section{Fast Separable Non-Local Means}
\label{sec: fast separable}

In the previous section, we considered the NLM of one-dimensional signals and presented a low-complexity algorithm called PatchLift-NLM for computing the same. 
We can, in principle, extend PatchLift to handle higher-dimensional patches. In particular, we could use the idea of lifting to compute the distance between patches drawn from an image. This would give us an algorithm for computing the NLM of images.
However, we will not pursue this direction in this paper, and instead we will take a different route by which we can achieve better denoising and a smaller run time. 

In particular, we propose a separable formulation of \eqref{NLM} in which we process the rows and the columns of the input image using \eqref{1dNLM}. 
We remark that we are not suggesting that  \eqref{NLM} can be computed by applying \eqref{1dNLM} along rows and columns. 
Indeed, the kernel in \eqref{NLM} is not separable and hence the NLM of an image cannot be computed by applying row and column operations.
By a separable formulation, we mean that we simply choose to process the rows using \eqref{1dNLM} and then process the columns of the intermediate image using \eqref{1dNLM}. The problem, however, is that the row and column operations do not commute, that is,  different results are obtained depending on whether the row are processed first or the columns. There are two distinct possibilities: 
\begin{itemize}
\item RC: we first filter the rows and then the columns of the intermediate image,
\item CR: we first filter the columns and then  the rows of the intermediate image.
\end{itemize}
We have generally noticed that by the final output obtained by averaging the images obtained using (RC) and (CR) operations exhibits a better visual appearance and a higher PSNR.  
More generally, we could linearly combine the two outputs and take it to be the final denoised image. 
A schematic of the proposal is provided in Figure \ref{schematic}. Along one pipeline, we first filter the rows of the noisy image $f$ using  \eqref{1dNLM}  to get $\fr$, and then we filter  the columns of $\fr$ using  \eqref{1dNLM}  to get $\frc$. 
In the other pipeline, we first filter the rows of $f$ using  \eqref{1dNLM} to get $\fc$, and then we filter the rows of $\fc$ using  \eqref{1dNLM} to get $\fcr$. 
We then linearly combine $\frc$ and $\fcr$ to get the final denoised image:
\begin{equation}
\label{snlm}
\tilde{f}(i)=\theta_1 \frc(i) + \theta_2 \fcr(i) \qquad (i \in \Omega),
\end{equation}
where we recall that $\Omega  = \big\{ i=(i_1, i_2):  1\leq i_1,i_2 \leq N\}$. 

\subsection{Optimal Combination}

The important question that we must address is how do we fix $\theta_1$ and $\theta_2$ in \eqref{snlm}? Moreover, it is important that the process of computing these parameters should be efficient; else, it will undermine the objective of having a fast algorithm.

In this regard, we suppose that the noisy image $f$ is derived from a clean image $f_0$ as follows:
\begin{equation}
\label{noise}
f(i) = f_0(i) + \sigma \cdot w(i),
\end{equation}
where $\sigma$ is the additive noise level, and $\{w(i) : i \in \Omega \}$ are independently drawn from $\mathcal{N}(0,1)$ \cite{Buades2005,KSVD,BM3D}. In keeping with \eqref{noise}, the parameters $\theta_1$ and $\theta_2$ should ideally be tuned to make $\tilde{f}$ ``similar'' to $f_0$. In particular, we can consider the mean-squared-error (MSE) given by
\begin{equation}
\label{mse}
\text{MSE} = \frac{1}{N^2} \sum_{i \in \Omega} \big(\tilde{f}(i) - f_0(i) \big)^2,
\end{equation}
and select $\theta_1$ and $\theta_2$ that minimizes \eqref{mse}. This appears to be an interesting proposition since it can be seen from \eqref{snlm} and \eqref{mse} that MSE is quadratic in the parameter vector $\boldtheta=(\theta_1,\theta_2)$. Thus, the optimal $\boldtheta$ can be obtained simply by setting the gradient of MSE to zero, and solving the resulting $2 \times 2$ linear system. Of course, the problem is that we do not have access to the clean image $f_0$ and hence we cannot compute \eqref{mse}.

 \begin{figure}
\centering
\includegraphics[width=0.6\linewidth]{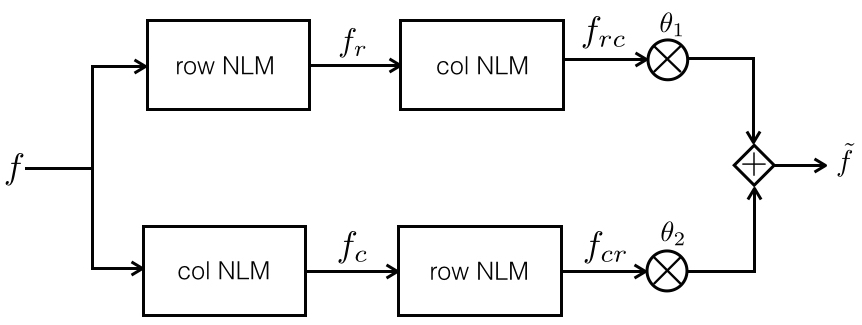}
\caption{Schematic of the proposed separable formulation of NLM. } 
\label{schematic}
\end{figure}

The remarkable fact is that we circumvent the above problem of not having the clean image using Stein's Unbiased Risk Estimator (SURE) \cite{S1981}. SURE was originally proposed in \cite{S1981} for estimating the mean of an isotropic Gaussian distribution from its iid samples. The fact that this is precisely the data model in \eqref{noise}, and that one can use SURE (as a surrogate of MSE) for transform-based image denoising, have been used in a series of papers; see, e.g., \cite{BL2007,VK2009}. We next explain how SURE can be used for optimizing the parameters in \eqref{snlm}.

The first observation is that the transformation that takes the  $f(i)$ into  $\tilde{f}(i)$ is differentiable. This is simply because \eqref{snlm} is obtained through rowwise and columnwise applications of \eqref{1dNLM}, which is clearly differentiable. In particular, the SURE of \eqref{1dNLM} is given by
\begin{equation}
\label{sure}
\phi= \frac{1}{N^2} \sum_{i \in \Omega} \big(\tilde{f}(i) - f(i) \big)^2 - \sigma^2 + \frac{2\sigma^2}{N^2} \sum_{i \in \Omega} \frac{\partial \tilde{f}(i)}{\partial f(i)}.
\end{equation}
The random variable $\phi$ is an unbiased estimator of MSE in the sense that its mathematical expectation (with respect to the underlying Gaussian noise distribution) is identical to the expectation of the MSE in \eqref{mse}. The proof of this fact comes as a direct consequence of the observations in the original paper \cite{S1981}. In other words, we can use instances of $\phi$ as a surrogate of MSE. In fact, due to the large size of images and the law of large numbers, Stein's estimator turns out to be a very stable surrogate of the mean-squared-error in practice \cite{BL2007,VK2009}.

Notice that \eqref{sure} depends on the noisy image and the divergence of the input-output mapping depicted in Figure \ref{schematic}. In particular, similar to the observation made earlier for MSE, the first term in \eqref{sure} is quadratic in $\boldtheta$. On the other hand, it follows from \eqref{snlm} that
\begin{equation}
\label{linear}
\frac{\partial \tilde{f}(i)}{\partial f(i)} =  \theta_1\frac{\partial \frc(i)}{\partial f(i)} + \theta_2 \frac{\partial \fcr(i)}{\partial f(i)}.
\end{equation}
Thus, the last term in \eqref{sure} is linear in $\theta$. Therefore, by substituting \eqref{snlm} and \eqref{linear} in \eqref{sure}, we get a quadratic function in $\boldtheta$. The optimum $\boldtheta$ that minimizes $\phi$ be obtained by setting the gradient (of $\phi$ with respect to $\boldtheta$) to zero. In particular, one can verify that the optimal assignment $\boldtheta^{\ast}$ satisfies 
\begin{equation*}
\mathbf{A}\boldtheta^{\ast} =\mathbf{b},
\end{equation*}
where
\begin{equation}
\label{A}
\begin{aligned}
\mathbf{A}=  \begin{bmatrix}
\  \underset{i \in \Omega}{\sum} \frc(i)^2   &&\underset{i \in \Omega}{\sum} \frc(i) \fcr(i) \\
\  \underset{i \in \Omega}{\sum}  \frc(i) \fcr(i)  &&\underset{i \in \Omega}{\sum} \fcr(i)^2
\end{bmatrix},
\end{aligned}   
\end{equation}
and
\begin{equation}
\label{b}
\begin{aligned}
\mathbf{b} = & \begin{bmatrix}
\  \underset{i \in \Omega}{\sum}f(i) \frc(i)   - \sigma^2 \underset{i \in \Omega}{\sum}\frac{\partial \frc(i)}{\partial f(i)}\\
\  \underset{i \in \Omega}{\sum}f(i) \fcr(i)   - \sigma^2 \underset{i \in \Omega}{\sum}\frac{\partial \fcr(i)}{\partial f(i)}
        \end{bmatrix}.
\end{aligned}
\end{equation}
In other words, we can minimize \eqref{sure} with the same computational ease with which we  could (hypothetically) minimize \eqref{mse}, namely, by solving a $2 \times 2$ linear system.

We now address the computation of the partial derivatives in \eqref{linear}, which appear in $\mathbf{b}$. It turns out that the separable formulation simplifies this computation. Indeed, since $\frc$ is obtained from $\fr$ using column filtering and $\fr$ is obtained from $f$ using row filtering, we can use the chain rule  to write
\begin{equation}
\label{partialrc}
\frac{\partial \frc(i)}{\partial f(i)} = \frac{\partial \frc(i)}{\partial \fr(i)} \cdot \frac{\partial \fr(i)}{\partial f(i)}.
\end{equation}
Similarly, we can write
\begin{equation}
\label{partialcr}
\frac{\partial \fcr(i)}{\partial f(i)} = \frac{\partial \fcr(i)}{\partial \fc(i)} \cdot \frac{\partial \fc(i)}{\partial f(i)}.
\end{equation}
However, notice that the partial derivatives on the right in \eqref{partialrc} and \eqref{partialcr} are simply the partial derivatives of the transformation in \eqref{1dNLM}, namely,
\begin{equation}
\label{partial}
\frac{\partial \hat{f}(i)}{\partial f(i)}  \qquad (1 \leq i \leq N) .
\end{equation}
In fact, the partial derivatives of the original two-dimensional NLM (with box kernel) was first derived in \cite{VK2009}. This formula is structurally identical to that for the one-dimensional NLM. In particular, for $1 \leq i \leq N$,
\begin{align}
\label{div}
&\frac{\partial \hat{f}(i)}{\partial f(i)}  =  \frac{2}{h^2 W_i}\sum_{j =i-S}^{i+S} w_{ij} f(j)^2 + \frac{1}{W_i} - \frac{2 }{h^2} \hat{f}(i)^2  \\
 &+  \frac{2}{W_i h^2}\!\! \sum_{k=i-K}^{i+K} \!\! w_{ik} g_{\beta}(i-k) \big(f(k) - \hat{f}(i)\big) \big(f(2i-k) - f(i)\big), \nonumber
 \end{align}
where $W_i = \sum_{j = i-S}^{i+S} w_{ij}$.
For completeness, the main steps leading to \eqref{div} are provided in the Appendix.

\LinesNumberedHidden
\begin{algorithm}
\KwData{Signal $f(1),\ldots,f(N)$, parameters $K,S,h$,  and 
a sequence $\delta_{\mathrm{in}}(1),\ldots,\delta_{\mathrm{in}}(N)$.}
\KwResult{NLM $\hat{f}$ given by \eqref{1dNLM}, and  $\delta_{\mathrm{out}}$ given by \eqref{div}.}
Compute  $\overline{V}$ from $f,K$, and $S$ using Algorithm \ref{algo1}\;
\For{$i=1,\ldots,N$} {
Set  $T_1, T_2, T_3, T_4 , P_1, P_2,$ and $Q$ to zero\;
\For{$j =  i - S,\ldots,i-K-1$}{
Compute $d_{ij}^2$ using \eqref{dist}\;
$w = \exp(-d_{ij}^2/h^2)$\;
$P_1=P_1+w  f(j)$\;
$P_2=P_2+w  f(j)^2$\;
$Q=Q+w$\;
}
\For{$j =  i - K,\ldots,i+K$}{
Compute $d_{ij}^2$ using \eqref{dist}\;
$w = \exp(-d_{ij}^2/h^2)$\;
$P_1=P_1+w  f(j)$\;
$P_2=P_2+w  f(j)^2$\;
$Q=Q+w$\;
$T_1 = T_1 + w  g_{\beta}(i - j) $\;
$T_2  =T_2 + w g_{\beta}(i - j) f(j)$\;
$T_3 = T_3 +  w g_{\beta}(i - j) f(2i-j) $\;
$T_4 = T_4 +  w g_{\beta}(i - j) f(2i-j) f(j)$\;
}
\For{$j =  i+K+1,\ldots,i+S$ }{
Compute $d_{ij}^2$ using \eqref{dist}\;
$w = \exp(-d_{ij}^2/h^2)$\;
$P_1=P_1+w  f(j)$\;
$P_2=P_2+w  f(j)^2$\;
$Q=Q+w$\;
}
$\hat{f}(i) =P_1/Q$\;
$T = T_1 \hat{f}(i) f(i) - T_2 f(i) -  T_3 \hat{f}(i)+ T_4$\;
$\delta_{\mathrm{out}}(i) = \delta_{\mathrm{in}}(i) (h^2+2T+2 ( P_2 - Q\hat{f}(i)^2)/Qh^2 $\;
} 
\caption{Computation of \eqref{1dNLM} and \eqref{div}.}
\label{algo3}
\end{algorithm}

\IncMargin{1.5mm}
\begin{algorithm}
\LinesNumbered
\KwData{Image $f$ of size $N \times N$, and parameters $K,S,h$.}
\KwResult{Separable NLM output $\tilde{f}$.}
\textbf{Initialize}: $\delta_1$ and $\delta_2$ of size $N \times N$ with entries set to $1$\;
\For{$i_1=1,\ldots,N$} {
\hspace{-3mm}$(\fr(i_1, :), \delta_1(i_1,:)) = \text{Alg3}(f(i_1, :), \delta_1(i_1,:), K, S, h)$
}
\For{$i_2=1,\ldots,N$} {
\hspace{-4mm} $(\frc(:, i_2), \delta_1(:,i_2)) = \text{Alg3}(\fr(:, i_2), \delta_1(:, i_2), K, S, h)$ \label{delta1}
}
\For{$i_2=1,\ldots,N$} {
\hspace{-4mm} $(\fc(:, i_2), \delta_2(:,i_2)) = \text{Alg3}(f(:, i_2), \delta_2(:, i_2), K, S, h)$
}
\For{$i_1=1,\ldots,N$} {
\hspace{-3mm}$(\fcr(i_1, :), \delta_2(i_1,:)) = \text{Alg3}(\fc(i_1, :), \delta_2(i_1,:), K, S, h)$\label{delta2}
}
Compute $\mathbf{A}$ and $\mathbf{b}$ using \eqref{A} and \eqref{bb}\;
Solve $\mathbf{A} \boldtheta^\star = \mathbf{b}$\;
Set $\tilde{f}(i)=\theta^{\ast}_1 \frc(i) + \theta^{\ast}_2 \fcr(i)$.
\caption{SNLM.}
\label{algo4}
\end{algorithm}
\DecMargin{1.5mm}

In summary, we have to perform additional $O(K)$ computations to determine the partial derivatives in \eqref{partial}.
We note that the computation of the terms in \eqref{div} can be integrated into PatchLift-NLM. This is summarized in Algorithm \ref{algo3}. The input to Algorithm \ref{algo3} is a one-dimensional signal, and we output its NLM and the partial derivatives in \eqref{partial}. The NLM is computed using Algorithm \ref{algo2}.

The complete algorithm for computing \eqref{snlm} along with the optimal $\boldtheta^{\ast}$ is summarized in Algorithm \ref{algo4}. We use Algorithm \ref{algo3} as a black-box (referred to as Alg3 during the call) in Algorithm \ref{algo4}. Notice that we input a sequence $\delta_{\mathrm{in}}$ of length $N$ in Algorithm \ref{algo3}, which is used for performing the chaining in \eqref{partialrc} and \eqref{partialcr} in Algorithm \ref{algo4}. We use the notations $f(i,:)$ and $f(:,j)$ to respectively denote the $i$-th row and the $j$-th column of an image $f$. The outputs $\delta_1$ and $\delta_2$ in steps \ref{delta1} and \ref{delta2} are respectively the partial derivatives in \eqref{partialrc} and \eqref{partialcr}. That is, 
\begin{equation}
\label{bb}
\begin{aligned}
\mathbf{b} = & \begin{bmatrix}
\  \underset{i \in \Omega}{\sum}f(i) \frc(i)   - \sigma^2 \underset{i \in \Omega}{\sum}\delta_1(i)\\
\  \underset{i \in \Omega}{\sum}f(i) \fcr(i)   - \sigma^2 \underset{i \in \Omega}{\sum}\delta_2(i)
        \end{bmatrix}.
\end{aligned}
\end{equation}
It follows from Corollary \ref{coro} and the above discussion that the overall complexity of Algorithm \ref{algo4} is $O(N^2S)$.

\begin{figure*}[htp!]
\centering
\subfloat[Clean.]{\includegraphics[width=0.24\linewidth]{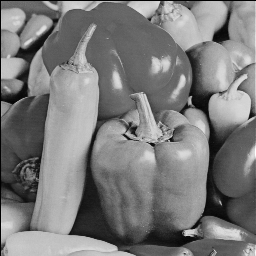}} \hspace{-0.5mm}
\subfloat[PSNR = 24.61 dB.]{\includegraphics[width=0.24\linewidth]{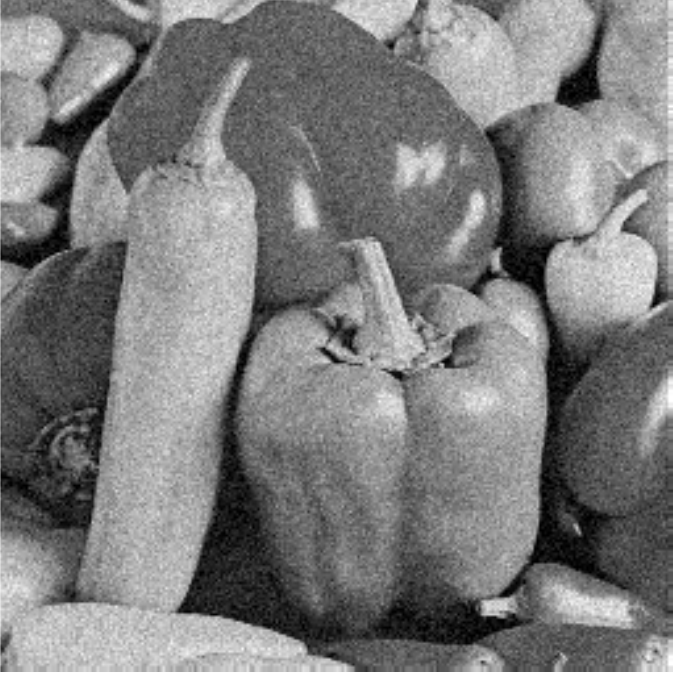}}  \hspace{-0.5mm}
\subfloat[PSNR = 28.27 dB.]{\includegraphics[width=0.24\linewidth]{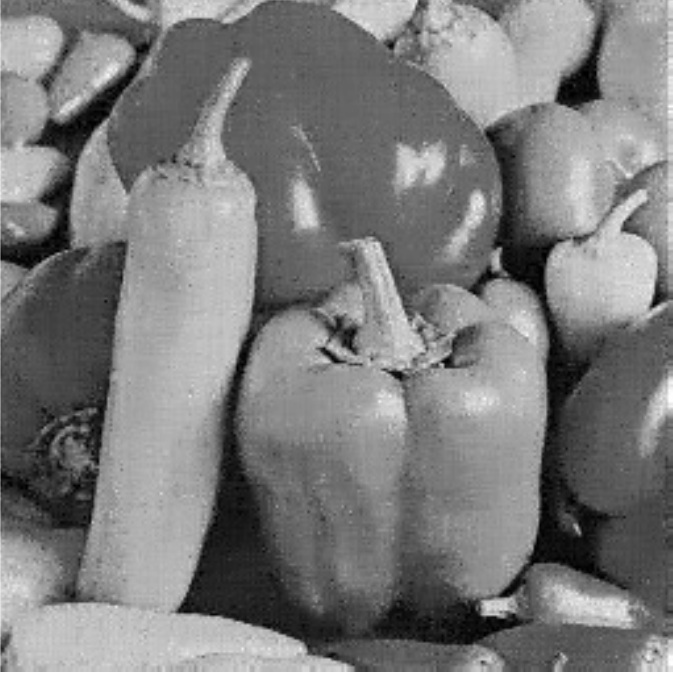}} \hspace{-0.5mm}
 \subfloat[PSNR = 31.52 dB.]{\includegraphics[width=0.24\linewidth]{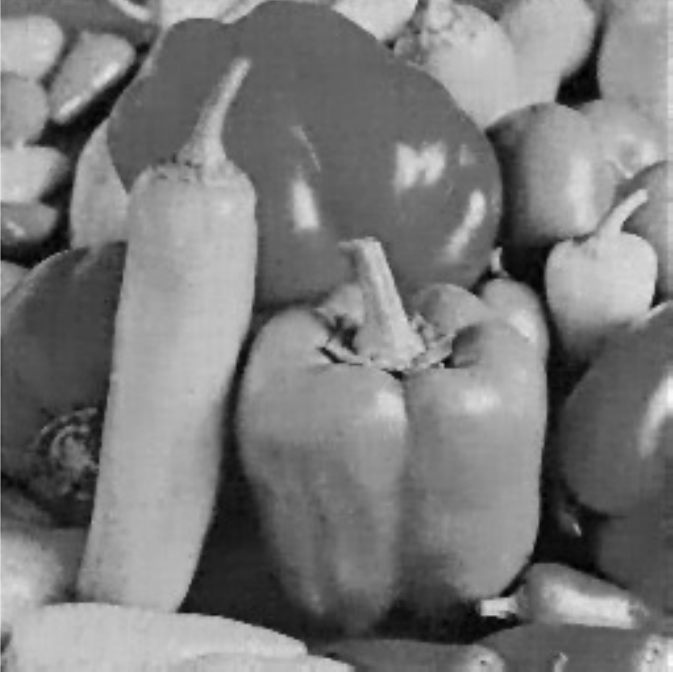}} 
\caption{Denoising of \emph{Peppers} using Algorithm \ref{algo4}, followed by bilateral filtering. 
(b) Noisy image ($\sigma=15$).
(c) Output of Algorithm \ref{algo3} obtained using a \textbf{box} kernel, where $h = 2.1 \sigma, K = 3$, and $S=10$.
(d) Bilateral filtering of (c) using $\sigma_s = 0.71$ and $\sigma_r = 77$. Notice that the grains appearing in (c) are diffused out after the bilateral filtering.}
\label{fig1}
\end{figure*}

\begin{figure*}[htp!]
\centering
\subfloat[Clean.]{\includegraphics[width=0.24\linewidth]{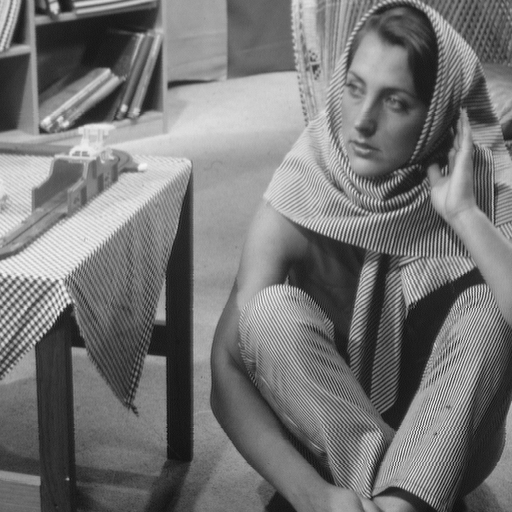}} \hspace{-0.5mm}
\subfloat[PSNR = 22.11 dB.]{\includegraphics[width=0.24\linewidth]{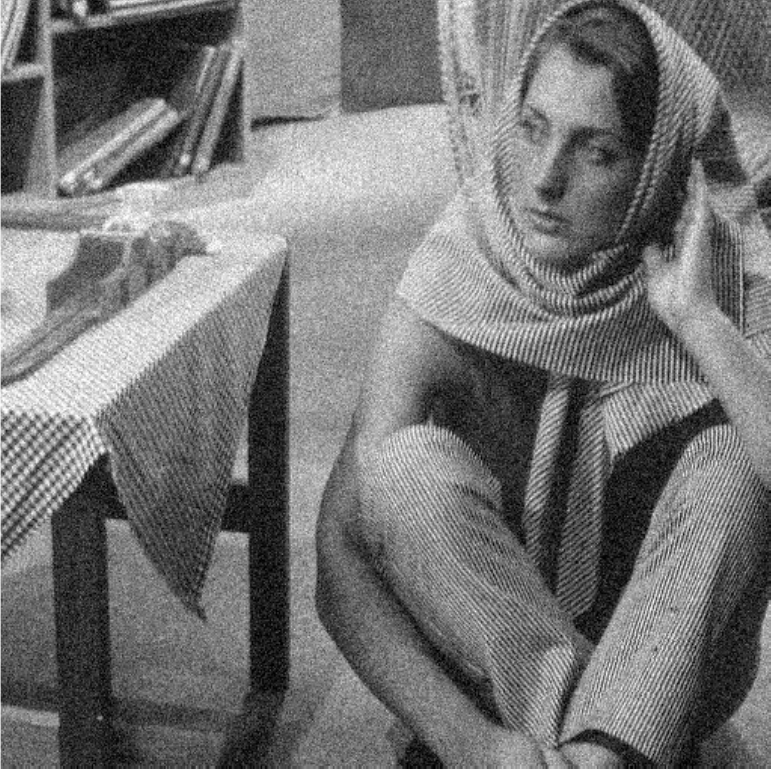}}\hspace{-0.5mm}
\subfloat[PSNR = 27.21 dB.]{\includegraphics[width=0.24\linewidth]{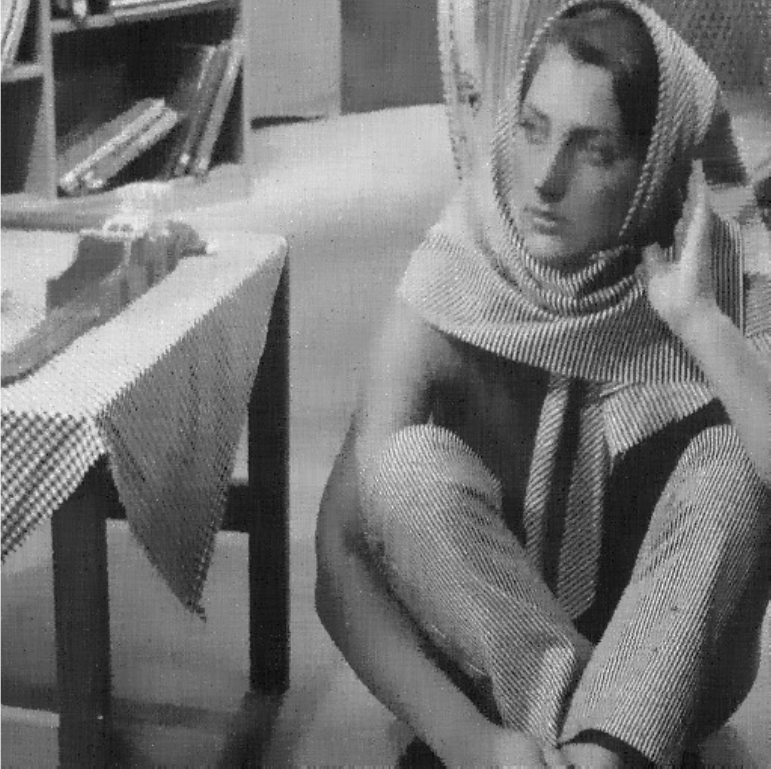}} \hspace{-0.5mm}
\subfloat[PSNR = 27.95 dB.]{\includegraphics[width=0.24\linewidth]{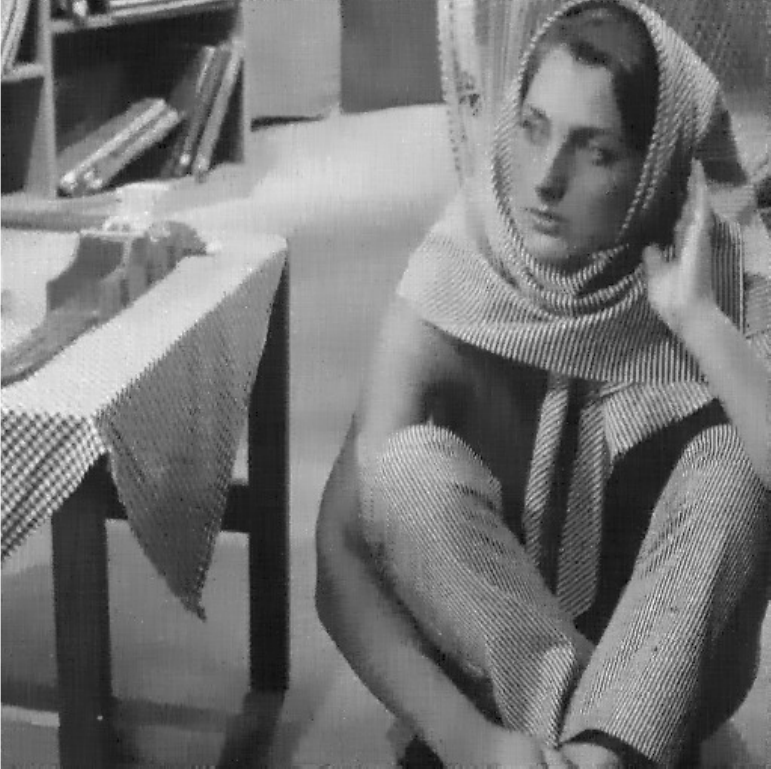}}
\caption{Denoising of \emph{Barbara} using Algorithm \ref{algo4}, followed by bilateral filtering. The difference with Figure \ref{fig1} is that we use a Gaussian kernel in this case.
(b) Noisy image ($\sigma=20$). (c) Output of Algorithm \ref{algo3} obtained using a \textbf{Gaussian} kernel, where $h = 2.1 \sigma, K = 3$, and $S=10$.
(d) Bilateral filtering of (c) using $\sigma_s = 0.76$ and $\sigma_r = 103$.}
\label{fig2}
\end{figure*}

\subsection{Post-Processing}

We have noticed that the separable processing introduces stripes into the final processed image. 
However, these stripes are of small amplitude, and extensive experiments reveal that these stripes can be suppressed by post-processing the output of Algorithm \ref{algo4} using a smoothing  filter.
In particular, we have used the non-linear bilateral filter \cite{Tomasi1998} to post-process the output of Algorithm \ref{algo4}.
The final denoised image $\text{SNLM}[f]$ is thus set to be
\begin{equation}
\label{BF}
\text{SNLM}[f](i)=  \frac{ \sum_{j \in \Omega} w(j)  g_{\sigma_r}(\tilde{f}(i-j)-\tilde{f}(i))  \tilde{f}(i-j)}{\sum_{j \in \Omega} w(j)  \  g_{\sigma_r}(\tilde{f}(i-j)-\tilde{f}(i)) } 
\end{equation}
where 
\begin{equation*}
w(i) = \exp\left(- \frac{\lVert i \rVert^2}{2\sigma_s^2}\right), \quad g_{\sigma_r}(t) = \exp\left(- \frac{t^2}{2\sigma_r^2}\right),
\end{equation*}
and $\Omega = [-3\sigma_s,3\sigma_s]^2$ is the support of the spatial Gaussian filter $w$. We will refer to the whole transformation that takes the input image $f$ into $\text{SNLM}[f]$ as Separable Non-Local Means. 

The edge-preserving action of the  bilateral filter diffuses out the stripes (and small grains), without compromising the structural information in the image.
We have implemented \eqref{BF} using the fast algorithm in \cite{Chaudhury2011}, whose complexity is $O(N^2)$ for an $N \times N$ image. 
We note that the cost of this bilateral filter variant does not scale with the
parameters of non-local means. Moreover, it does not scale with the $\sigma_s$ and $\sigma_r$ parameters of the
bilateral filter. 
Thus, the overall complexity of computing  $\text{SNLM}[f]$ is $O(N^2 S) + O(N^2 ) = O(N^2 S)$.
In practice, the overhead run time for the bilateral filtering is about $30\%$ of the run time of Algorithm \ref{algo4}. A visual comparison of the results before and after the application of the bilateral filter is provided in Figures \ref{fig1} and \ref{fig2}. Notice that the visual appearance and the denoising quality (as measured using $\text{PSNR}=10\log_{10}(255^2/\text{MSE})$) improves significantly after the post-processing.

\section{Experimental Results}
\label{sec:results}

\subsection{Parameters}

The parameters involved in SNLM are those required for the one-dimensional NLM ($\beta$ and $h$), and those required for the bilateral filtering ($\sigma_s$ and $\sigma_r$). In order to understand the sensitivity of various parameters and to learn the optimal rule for setting these parameters as a function of the noise level, we performed an exhaustive study using various natural images obtained from the repositories at \cite{BM3Dimages,SIPI,IPOL}. For a fixed image and noise level, we jointly optimized the parameters to get the maximum PSNR (for a fixed $K$ and $S$). 
The study revealed that the critical parameters are $h,\sigma_s,$ and $\sigma_r$.  For example, when $S=10$ and $K=3$, the optimal $h$ was found to be in the interval $[2.1\sigma, 2.6\sigma]$ for the box kernel, and in the interval $[1.8 \sigma, 2.3 \sigma]$ for the Gaussian kernel. The optimal rule for the bilateral filter was found to be
\begin{eqnarray*}
\begin{aligned}
 \sigma_s = & 2.5 \times 10^{-6} \sigma^3 - 3.4 \times 10^{-4}  \sigma^2 + 0.021 \sigma + 0.46, \text{ and} \\
 \sigma_r = & 2.8 \times 10^{-4}\sigma^3 - 0.088 \sigma^2 + 8 \sigma -24.
\end{aligned}
\end{eqnarray*}
For example, $\sigma_s \approx 1$ and $\sigma_r \approx 145$ when $\sigma=30$. The implementation of the bilateral filter in \cite{Chaudhury2011} is known to be fast for such large $\sigma_r$. The denoising performance is apparently not very sensitive to $\beta$, and $\beta=2$ was near-optimal for most images and noise levels. Similarly, we found $\alpha=2$ to be near-optimal for NLM. We note that we can alternatively use SURE to optimize the parameters, which was the original proposal in \cite{VK2009} for NLM. This would, of course, require us to run the algorithm multiple times (which is reasonable for a fast algorithm).

\subsection{Run-Time}

We now present simulation results concerning the run-time of the proposed algorithm. The computations were performed using a 3.40 GHz Intel 8-core machine with 32 GB memory. We also compared the proposed algorithm with the fast algorithms listed in Table \ref{complexity}. All algorithms were implemented using Matlab 8.4. For SNLM, we used the learnt parameter settings as described above. As for the rest of the algorithms, we use the settings suggested in the respective papers. 

\begin{figure}%{!htp}
\centering
\subfloat[$N=256$, $S = 20$.]{\includegraphics[width=0.4\linewidth]{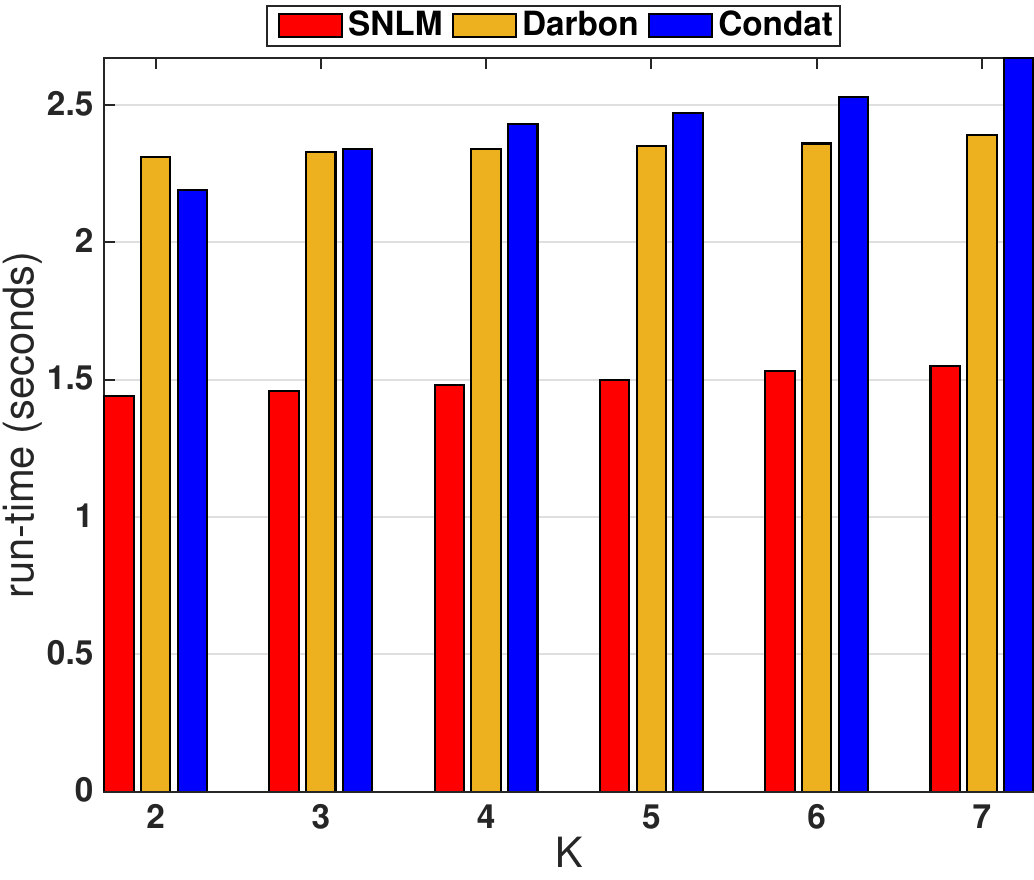}}  \hspace{0.5mm}
\subfloat[$N= 512$, $S = 20$.]{\includegraphics[width=0.4\linewidth]{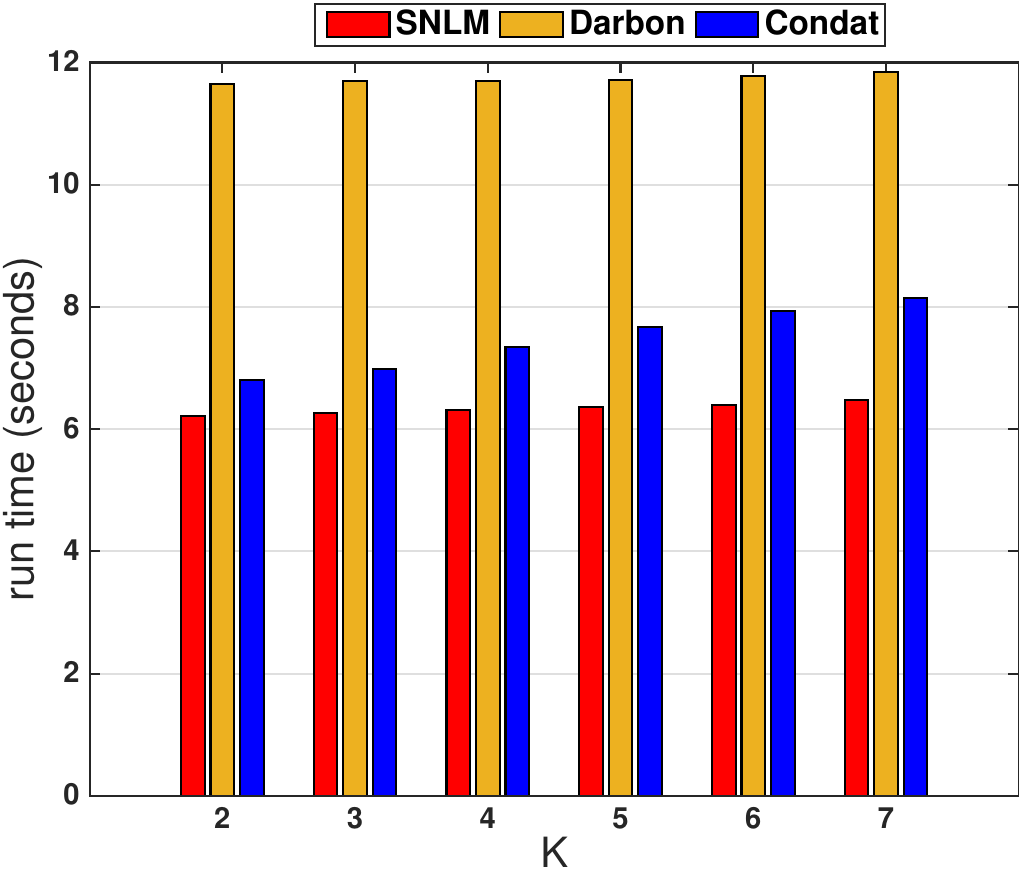}} \\ %\hspace{0.5mm}
\subfloat[$N=256$, $K = 3$.]{\includegraphics[width=0.4\linewidth]{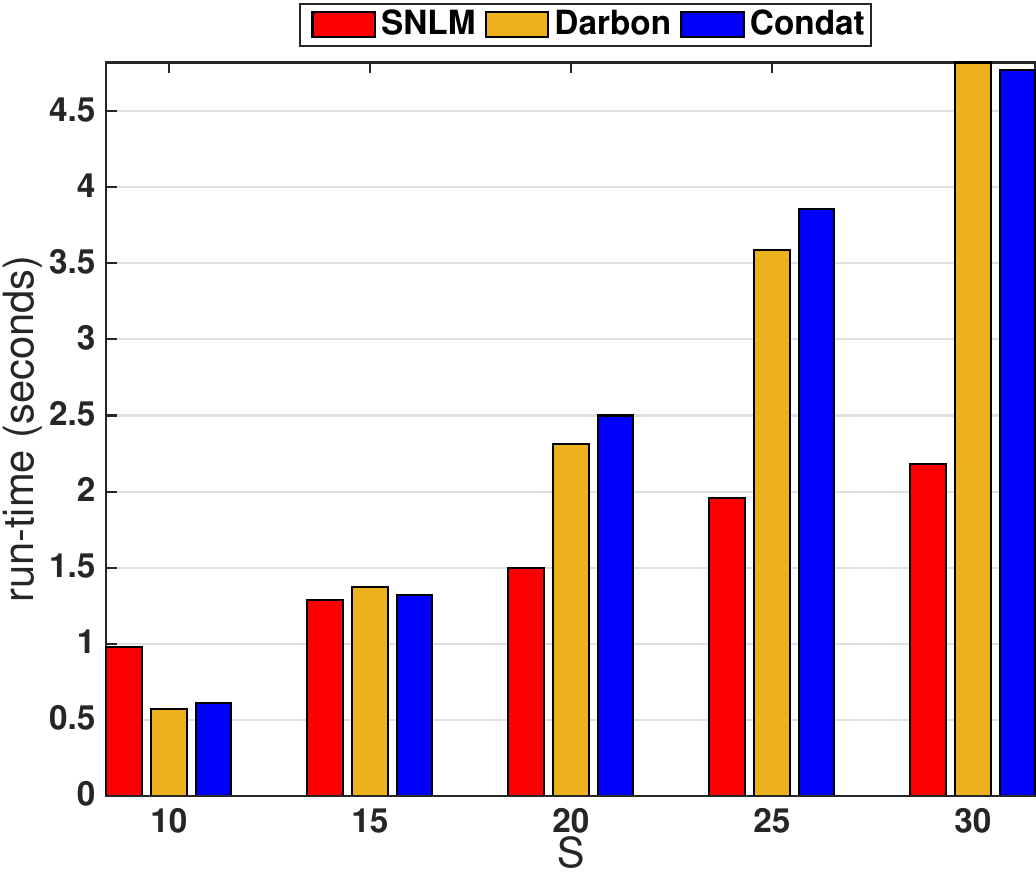}} \hspace{0.5mm}
\subfloat[$N= 1024$, $K = 3$.]{\includegraphics[width=0.4\linewidth]{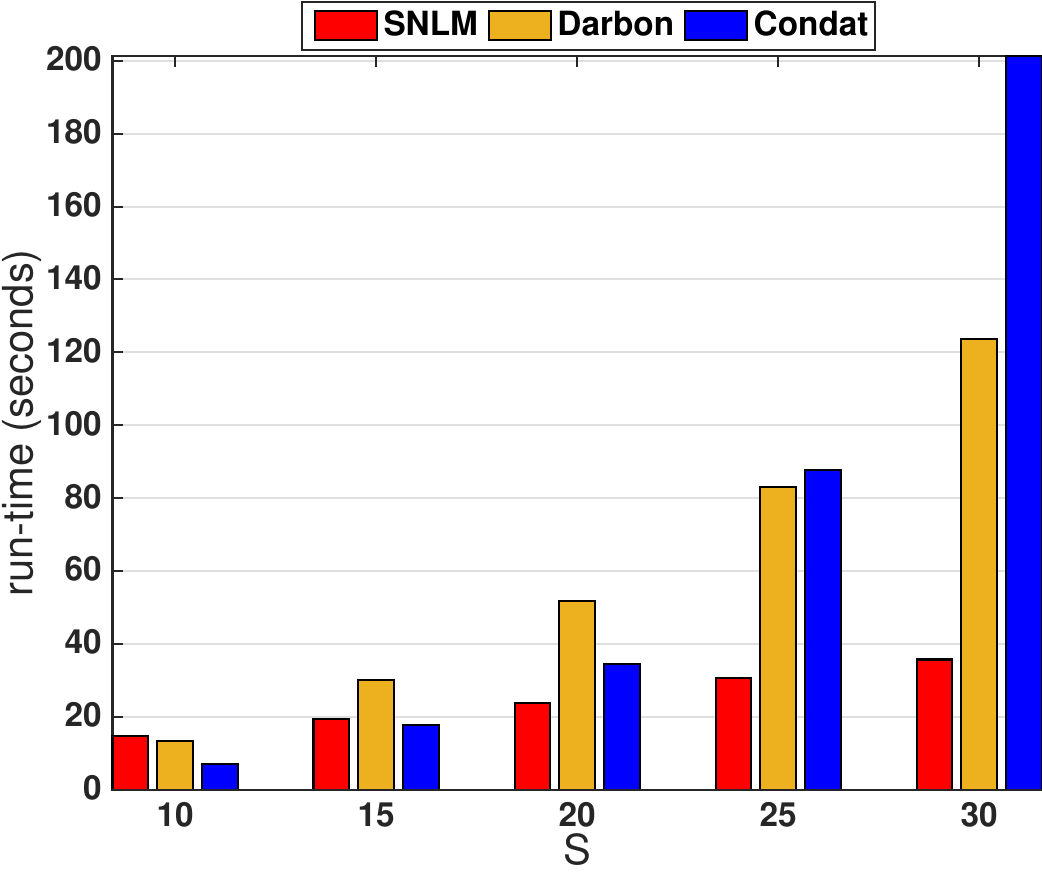}} 
\caption{Comparison of the run-time of the proposed SNLM with the fast algorithms from  \cite{Darbon2008} and \cite{Condat2010} for various image sizes (using a \textbf{box} kernel). Plots (a) and (b) are for a fixed $S = 20$ and $K \in \{2,3,\ldots,7\}$, while plots (c) and (d) correspond to $K = 3$ and $S  \in  \{10, 15,\ldots,30\}$.
} 
\label{barPlots}
\end{figure}

\begin{table}[htp!]
\setlength{\tabcolsep}{2.5pt}
\caption{Comparison of the run-time of Algorithm \ref{algo4} with NLM and its variants for various image sizes using a \textbf{box} kernel. The parameters are $S=20$ and $K=5$. The computations were performed using Matlab on a 3.40 GHz Intel 8-core machine with 32 GB memory.} 
\vspace{2mm}
\centering 
\begin{tabular}{|c| c| c | c | c | c | c|}
\hline
N       & NLM \cite{Buades2005}      & PCA  \cite{Tasdizen2009}    & Wang  \cite{Wang2006}   & Darbon \cite{Darbon2008}  & Condat  \cite{Condat2010} & SNLM  \\
\hline
$256$       & 497s     & 9.70s   & 25.24s   & 2.35s    & 2.47s    & {\bf 1.50}s \\ % 2005
\hline
$512$      & 34m      & 43.4s    & 104.3s   & 11.78s   &7.67s     & {\bf 6.32}s  \\ %ICIP 2011
\hline
$1024$     & 136m     & 165.8s     & 430s     &53.31s    & 89.18s   &  {\bf28.77}s  \\ % ICIP 2006
\hline 
\end{tabular}
\label{timingBox}
\end{table}

\begin{table}[htp!]
\setlength{\tabcolsep}{3.0pt}
\caption{The setting in this case is identical to that used in Table \ref{timingBox}, except that we have used a \textbf{Gaussian} kernel instead of a box.}
\vspace{2mm}
\centering 
\begin{tabular}{|c| c | c | c | c |c |}
\hline
N          & NLM  \cite{Buades2005}    & Mahmoudi \cite{Sapiro2005}  & Brox \cite{Brox2008}   & Vignesh \cite{earlytermination2010}    & SNLM
 \\ \hline
$256$       & 512s     &245s       & 184s    & 92s       & {\bf1.7}s
\\
\hline
$512$     &  38m     & 15.2m     & 13.0m   & 6.46m      & {\bf6.8}s
\\
\hline
$1024$    & 145m     & 52.2m     & 46.4m   &20.3m       & {\bf 30.4}s
\\
\hline 
\end{tabular}
\label{timingGaussian}
\end{table}

It is clear from the complexity estimates in Table \ref{complexity} that the scaling with respect to $S$ is minimal for SNLM (and there is no scaling with $K$). This is evident from the results reported in Tables \ref{timingBox} and \ref{timingGaussian}, and  Figure \ref{barPlots}. In Figure \ref{barPlots}, we have compared SNLM with the fast algorithms in \cite{Darbon2008} and \cite{Condat2010}. The reason is that we found these to be the leading algorithms in terms of run-time.  We notice from the plots in Figure \ref{barPlots} that SNLM runs much faster for large $S$ and $K$. Since $S$ and $K$ typically scale with the image size, this means that SNLM outperforms \cite{Darbon2008} and \cite{Condat2010} for megapixel images. 

We have separately compared SNLM with NLM and other competing algorithms for box and Gaussian kernels in Tables  \ref{timingBox} and \ref{timingGaussian} (recall from Table \ref{complexity} that not all algorithms work with the Gaussian kernel). We notice that the PatchLift-based implementation of SNLM is significantly faster than the rest of the algorithms. In particular, as claimed in the Introduction, SNLM is at least $300$ times faster than NLM when $S=20$ and $K=5$.

\subsection{Denoising Results}

Finally, we present some results concerning the denoising performance of SNLM, and compare it with NLM and its variants. In this regards, we note that while some of the fast algorithms mentioned above can be used for implementing the original NLM from \cite{Buades2005}, the authors of these algorithms have proposed various modifications that can lead to a better denoising performance. We have incorporated these improvements in the implementation; as a result, the PSNR obtained using these algorithms are generally different from that obtained using NLM. We have also used the structural-similarity index (SSIM) for evaluating the denoising performance \cite{SSIM2004}. 

\begin{figure}[htp!]
\centering
\subfloat[Clean.]{\includegraphics[width=0.4 \linewidth]{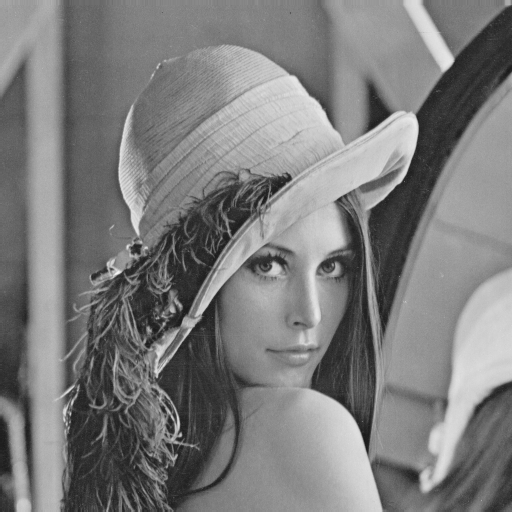}} \hspace{0.5mm}
\subfloat[Noisy (22.12, 67.95).]{\includegraphics[width=0.4 \linewidth]{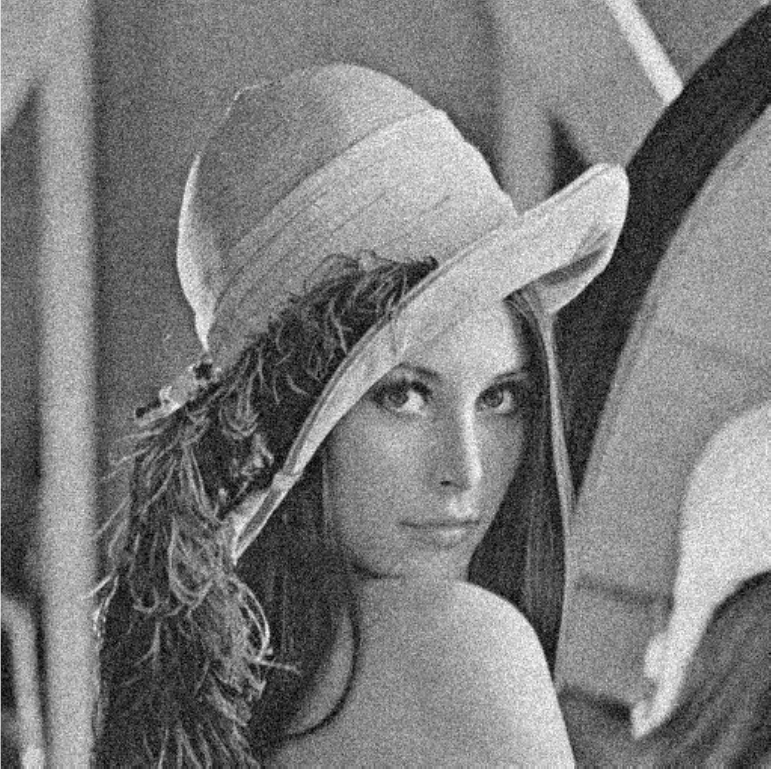}} \\ \vspace{-0.3cm}
\subfloat[NLM \cite{Buades2005} (30.46, 89.17).]{\includegraphics[width=0.4 \linewidth]{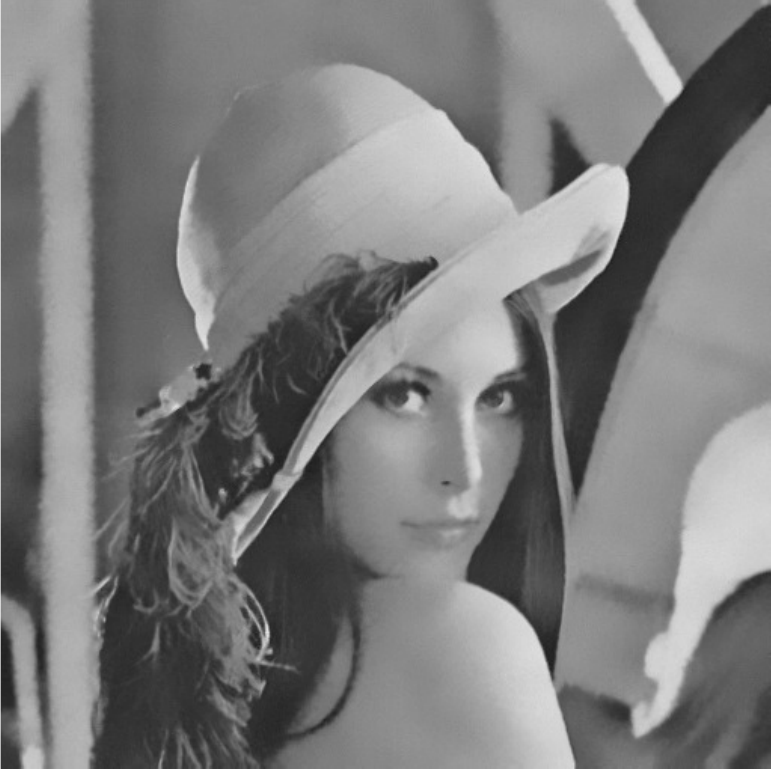}} \hspace{0.5mm}
\subfloat[SNLM (\textbf{31.44}, \textbf{91.38}).]{\includegraphics[width=0.4 \linewidth]{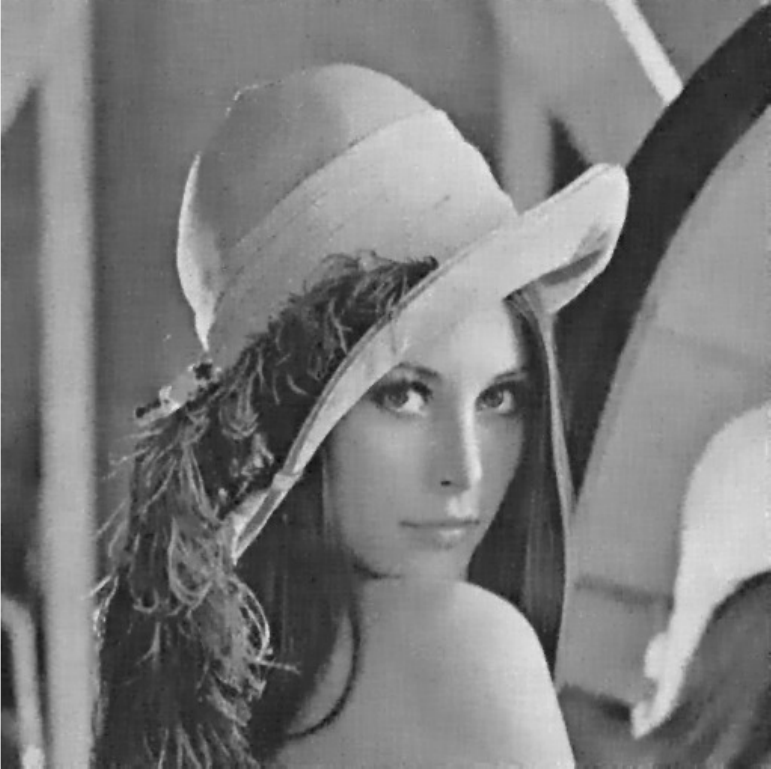}}  \\ \vspace{-0.3cm}
\subfloat[Method Noise for NLM.]{\includegraphics[width=0.4 \linewidth]{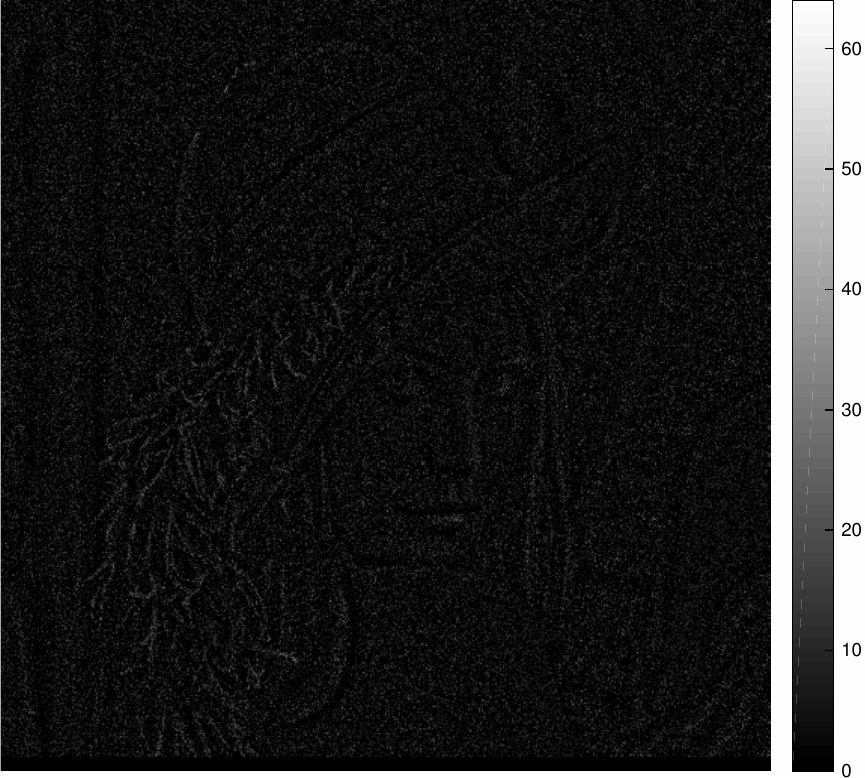}} \hspace{0.5mm}
\subfloat[Method Noise for SNLM.]{\includegraphics[width=0.4 \linewidth]{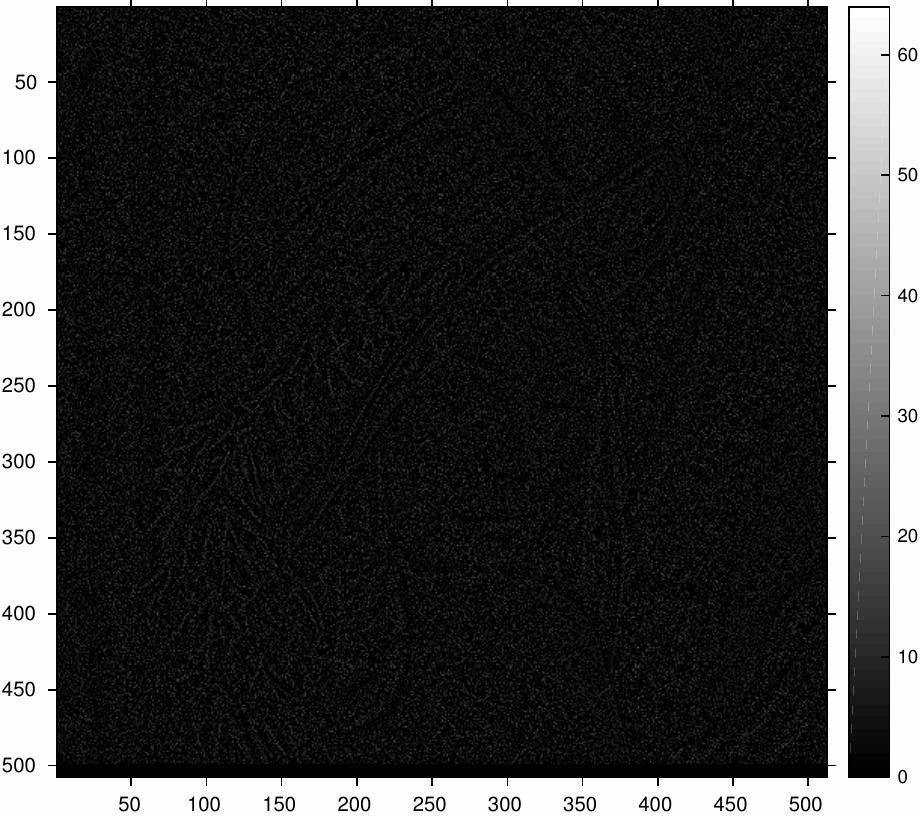}}
\caption{Denoising of \emph{Lena} ($512 \times 512$) at $\sigma=20$ using a \textbf{Gaussian} kernel for NLM ($\alpha=2$) and SNLM ($\beta=2$). Parameters: $S=10$ and $K=3$. The (PSNR, SSIM) indices are provided in the caption.}
\label{Lena}
\end{figure}

\begin{figure}[htp!]
\centering
\subfloat[Clean.]{\includegraphics[width=0.3\linewidth]{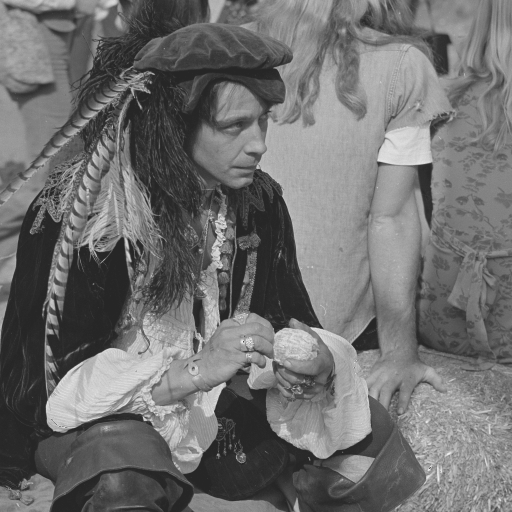}} \hspace{0.2mm}
\subfloat[Noisy (22.1, 74.1).]{\includegraphics[width=0.3 \linewidth]{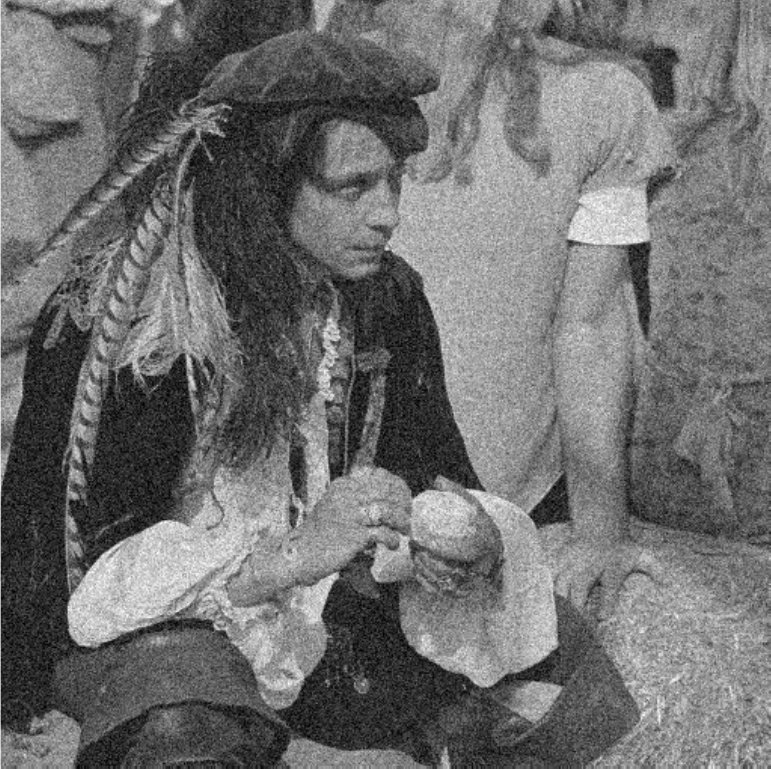}}  \\ \vspace{-0.3cm} 
\subfloat[NLM \cite{Buades2005} (26.8, 77.3).]{\includegraphics[width=0.3 \linewidth]{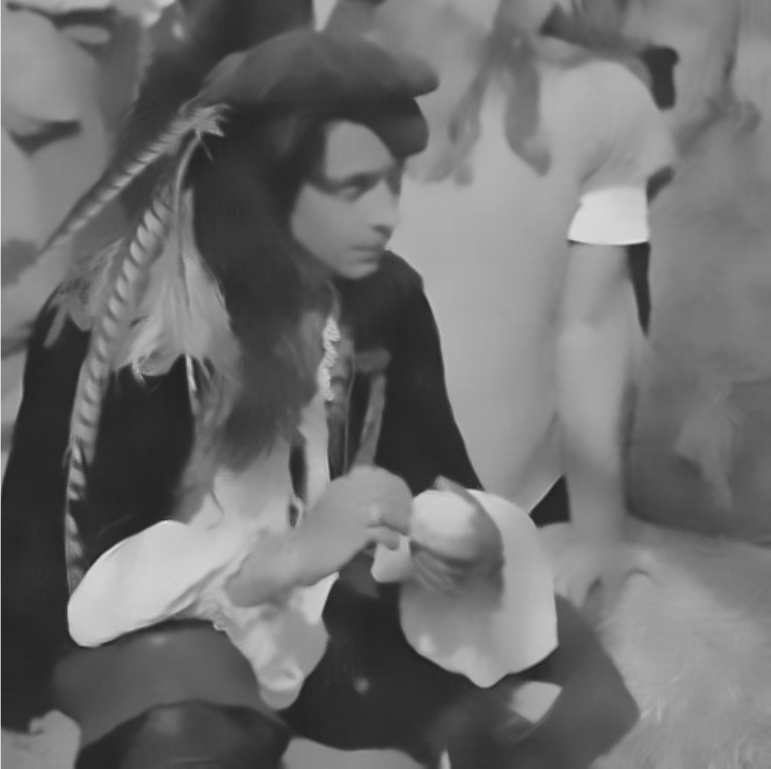}} \hspace{0.2mm}
\subfloat[SNLM (\textbf{29.6, 89.0}).]{\includegraphics[width=0.3\linewidth]{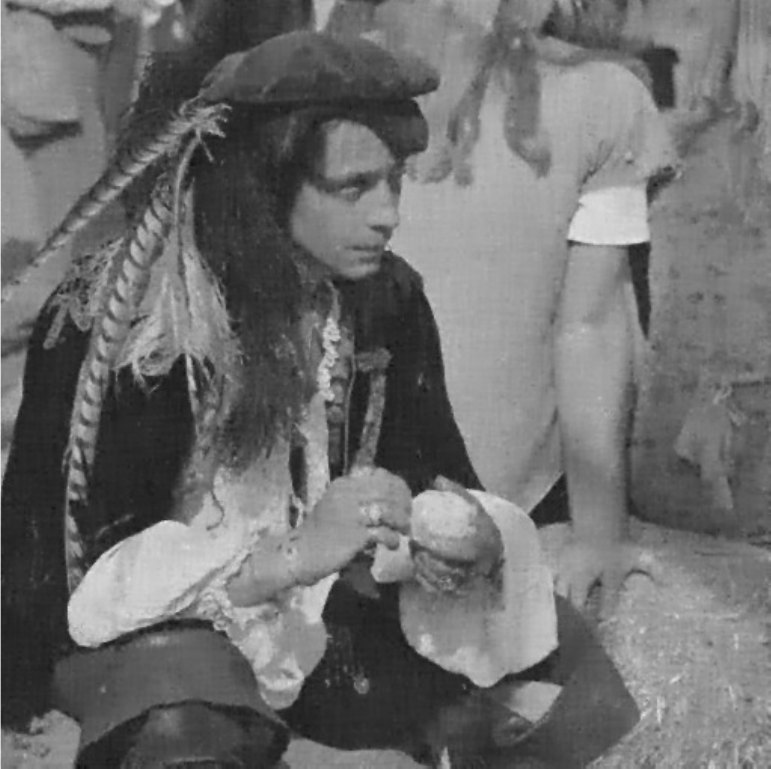}} \\ \vspace{-0.3cm} 
\subfloat[Method Noise for NLM.]{\includegraphics[width=0.3 \linewidth]{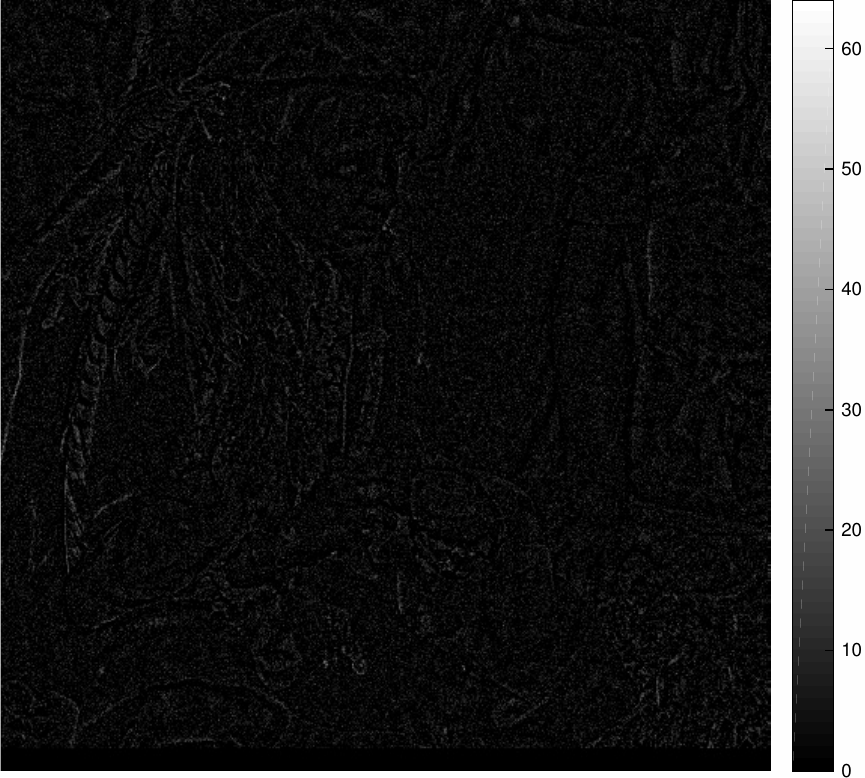}}  \hspace{0.2mm}
\subfloat[Method Noise for SNLM.]{\includegraphics[width=0.3 \linewidth]{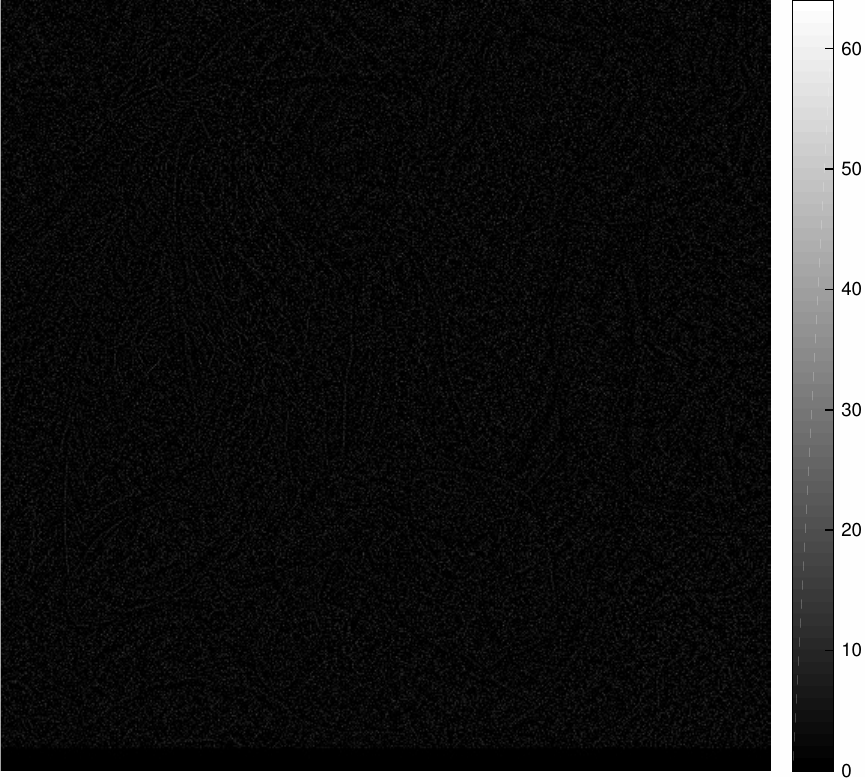}} \\ \vspace{-0.3cm} 
\subfloat[Darbon \cite{Darbon2008} (25.4, 83.2). ]{\includegraphics[width=0.3 \linewidth]{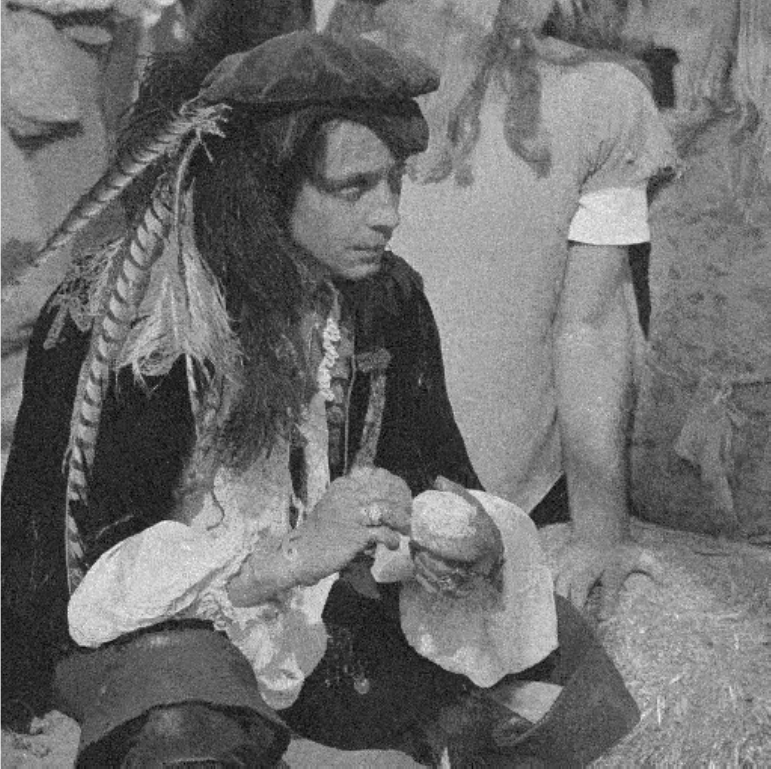}} \hspace{0.2mm} 
\subfloat[Condat \cite{Condat2010} (27.1, 79.1). ]{\includegraphics[width=0.3 \linewidth]{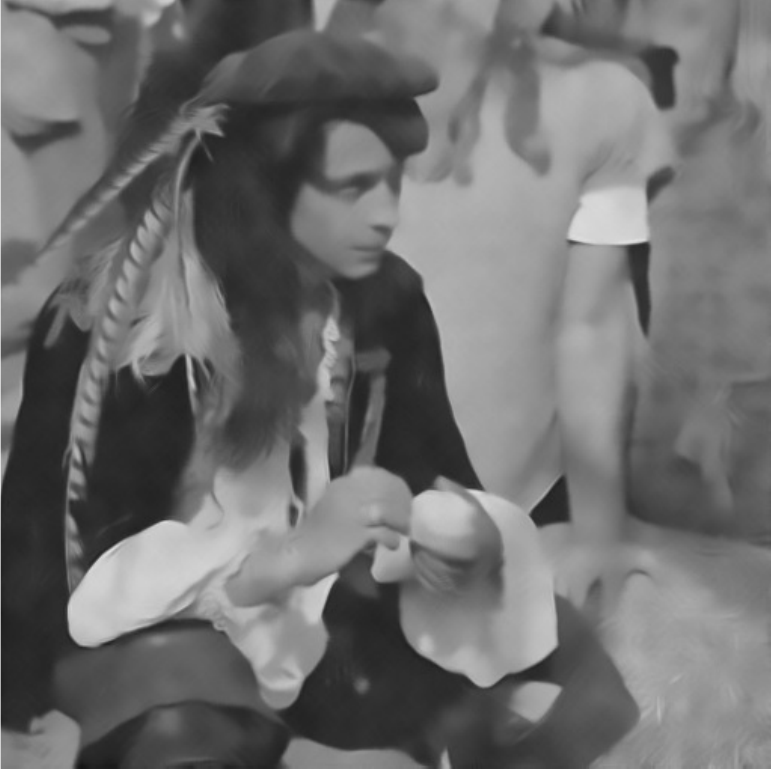}} 
\caption{Denoising of \emph{Man} (size $512 \times 512$) at $\sigma=20$ using a \textbf{box} kernel. Parameters: $S=10$ and $K=3$. The (PSNR, SSIM) indices are provided in the caption.} 
\label{Man}
\end{figure}

The denoising results on some standard natural images 
% (using box and Gaussian kernels) are provided in Tables \ref{psnrBox} and \ref{psnrGaussian}. 
(using Gaussian and box kernels) are provided in Tables \ref{psnrGaussian} and \ref{psnrBox}. 
We notice that the PSNR obtained using SNLM is consistently higher than that obtained using NLM; the gap is as large as $2$ dB at large noise levels. 
In fact, we are able to get better denoising by optimizing the parameters of our algorithm. As
is well-known, the $h = 10 \sigma$ rule for NLM is usually suboptimal. One can obtain better results
using NLM (comparable to that obtained using our method) by adjusting $h$ in NLM.
A visual comparison of the denoising quality is provided 
% in Figures \ref{Man} and \ref{Lena} for the box and Gaussian kernels. 
in Figures  \ref{Lena} and \ref{Man} for the Gaussian and box kernels. 
We notice that the output of SNLM is much more sharp compared to NLM and \cite{Condat2010}. Indeed, notice that in Figure \ref{Man}, the PSNR of the SNLM output is about $3$ dB higher than that of the NLM counterpart. Moreover, for both images, the method noise (difference of the noisy and the denoised image) \cite{BCM2010} obtained using SNLM exhibits less images structures than that obtained using NLM.

\begin{table*}[htp!]
\small{
\setlength{\tabcolsep}{3.0pt}
\centering 
\begin{tabular}{|c |c| c |c |c |c |c |c |}
\hline
Image &Method      & $\sigma = 10$  & $\sigma = 20$ & $\sigma = 30$	 & $\sigma = 40$  & $\sigma = 50$  
\\ \hline
& NLM    &  32.02/93.45  &  27.48/82.42  &  25.41/74.12  &  24.21/68.74  &  
23.41/63.71   
\\    \raisebox{0.2 cm}
{\emph{C2}}  % C2 = Couple
\raisebox{0.01 cm}
&SNLM   &  {\bf  32.74/96.18}  & {\bf29.14/90.49}  &  {\bf27.25/84.79}  &  {\bf25.96/79.33}  &
{\bf25.05/75.54}   
\\ 
\hline
& NLM   &  34.15/94.82  &  30.48/89.37  &  28.45/85.69  &  26.98/81.33  & 
25.87/77.56   
\\    \raisebox{0.2 cm}
{\emph{L1}} % L1 = Lena
\raisebox{0.01 cm}
&SNLM    &  {\bf34.64/96.11}  &  {\bf31.48/91.42}  &  {\bf29.59/87.92}  &  {\bf28.23/84.21}  & 
{\bf27.16/80.91}   
\\ 
\hline
& NLM   &  32.03/93.45  &  28.22/84.34  &  26.11/77.96  &  24.76/72.82  &  
23.86/68.09     
\\     \raisebox{0.2 cm}
{\emph{B3}} % B3  = Boat 
\raisebox{0.01 cm}
&SNLM    &  {\bf32.88/96.19}  &  {\bf29.47/90.07}  &  {\bf27.62/84.37}  &  {\bf26.35/80.15}  &  
{\bf25.39/75.40}   
\\ 
\hline
& NLM   &  31.78/91.25  &  28.19/81.89  &  26.43/73.94  &  25.35/68.71  &  
24.63/64.85     
\\     \raisebox{0.2 cm}
{\emph{H2}} % H2 = Hill 
\raisebox{0.01 cm}
&SNLM    &  {\bf32.88/95.31}  &  {\bf29.71/89.57}  &  {\bf28.01/82.11}  &  {\bf26.87/77.54}  & 
{\bf26.04/73.49}   
\\ 
\hline
\end{tabular}
\caption{Comparison of the denoising performance in terms of PSNR and 
SSIM at various noise levels using the \textbf{Gaussian} kernel with  
parameters $S = 10$ and $K=3$. The images used are \emph{C2: 
Couple}, \emph{L1: Lena}, \emph{B3: Boat}, and \emph{H2: Hill}. The 
size of each image is $512 \times 512$.}
\label{psnrGaussian}
}
\end{table*}

\begin{table*}[htp!]
\small{
\setlength{\tabcolsep}{3.0pt}
\centering 
\begin{tabular}{|c | c| c c c c c c |}
\hline
\small{Image} & $\sigma$   & NLM   & PCA & Wang & Darbon   & Condat  & SNLM 
\\ \hline
&10  & 30.72/88.36   & 32.34/94.88  & 30.72/88.36   & 30.37/93.80  & 30.19/88.34  & {\bf33.16/95.56}     \\
&20  & 26.87/77.30   & 28.91/88.78   & 26.87/77.30   & 25.46/83.18  & 27.16/79.13  & {\bf29.62/89.11}  \\ 
& 30 & 25.25/70.48   & 27.07/82.54   & 25.25/70.48  & 22.43/72.25  & 25.71/73.35  & {\bf27.79/83.41}  \\
\emph{M1}
&40 & 24.32/65.79   & 26.04/78.08  & 24.32/65.79  & 20.22/62.37  & 24.84/69.34  & {\bf26.55/78.40}   \\
&50 & 23.68/62.31   & 25.11/{\bf75.14}  & 23.68/62.31  & 18.47/54.06  & 24.21/66.40  & {\bf25.66}/74.37  \\ 
&80 & 22.58/55.86   & 22.80/60.30   & 22.58/55.86  & 14.70/36.40  & 22.88/60.20  & {\bf23.89/64.93} \\ 
\hline
&10 & 32.37/94.64   & 31.95/94.45  & 32.37/94.64  & 30.37/94.66  & 31.99/94.75  & {\bf 32.53/95.78}  \\  
&20  & 27.38/86.58   & {\bf28.30}/87.22   & 27.38/86.58  & 25.29/85.18  & 27.71/87.84  & 27.95/{\bf88.96}\\  
& 30 & 24.93/79.39   & 25.02/81.76   & 24.93/79.39  & 22.21/75.13  & 25.36/81.48  & {\bf25.15/83.08}  \\ 
\emph{B1}  
& 40 & 23.53/73.54   & 24.20/82.84   & 23.53/73.54 & 19.99/65.80  & 23.99/76.31  & {\bf24.09/78.15} \\ 
&50 & 22.64/68.88   & 23.21/71.09   & 22.64/68.88 & 18.26/57.69  & 23.13/72.35  & {\bf23.35/74.00} \\ 
&80 & 21.32/60.28   & 21.71/64.76   & 21.32/60.28  & 14.56/40.17  & 21.73/64.73  & {\bf 22.03/65.44} \\  
\hline
&10 & 34.22/86.89   & 34.52/{\bf89.16}  & 34.22/86.89  & 31.39/75.30  & 34.11/86.79  & {\bf34.83}/88.98 \\  
&20  & 29.75/81.80   & 30.80/83.07    & 29.75/81.80  & 26.09/50.70  & 30.17/82.03  & {\bf31.67/84.12}  \\  
 & 30 & 26.89/77.04   & 28.62/76.06  & 26.89/77.04  & 22.81/35.68  & 27.65/77.67  & {\bf29.70/80.17}\\ 
\emph{H1}
& 40 & 25.19/72.98   & 26.97/68.32  & 25.19/72.98   & 20.45/26.63  & 26.06/73.82  & {\bf28.17/76.58} \\ 
&50 & 24.12/69.67   & 25.76/62.05  & 24.12/69.67  & 18.61/20.71  & 24.98/70.51  & {\bf26.95/71.83} \\ 
&80  & 22.33/62.12   & 22.77/44.28   & 22.33/62.12  & 14.73/11.32  & 22.76/62.25  & {\bf24.37/66.20 }\\  
\hline
&10  & 28.07/82.34   & {\bf29.06/89.99}  & 28.07/82.34   & 27.57/87.83  & 24.16/65.78  & {28.88/88.83}  \\  
&20  & 22.17/50.50   & 24.60/{\bf76.33}  & 22.17/50.50  & 23.69/{73.35}  & 21.87/47.80  & {\bf24.64}/72.56 \\  
& 30 & 20.54/39.40   & 22.77/60.58    & 20.54/39.40  & 21.11/60.28  & 20.81/40.06  & {\bf 22.92/61.23}    \\ 
\emph{B2}
& 40 & 19.76/34.07   &  21.65/48.06    & 19.76/34.07  & 19.17/49.56  & 20.16/35.63  & {\bf 21.94/53.59}\\ 
&50 & 19.28/30.88   & 20.89/42.60   & 19.28/30.88  & 17.64/41.11  & 19.72/32.86  & {\bf 21.24/47.89} \\ 
&80 & 18.53/26.13   & 19.37/31.05 & 18.53/26.13 & 14.25/24.85  & 18.86/28.57  & {\bf19.88/37.46}  \\  
\hline
&10 & 31.31/86.51   & 32.38/90.90   & 31.31/86.51  & 30.35/76.99  & 29.79/85.32  & {\bf 33.15/91.85}   \\  
&20  & 27.88/80.37   & 28.27/82.71   & 27.88/80.37  & 25.57/54.03  & 27.28/80.02  & {\bf28.79/84.32}  \\  
& 30 & 24.78/74.97   & 26.03/75.07   & 24.78/74.97  & 22.46/40.03  & 24.90/75.26  & {\bf26.68/79.08}  \\ 
\emph{C1}
&40 & 22.94/70.34   & 24.61/68.30  & 22.94/70.34  & 20.17/31.34  & 23.34/70.96  & {\bf25.36/73.94}   \\ 
&50 & 21.86/66.54   & 23.48/61.00   & 21.86/66.54  & 18.38/25.49  & 22.30/67.05  & {\bf24.30/69.89}  \\ 
&80 & 20.16/58.03   & 21.14/43.19   & 20.16/58.03  & 14.55/15.42  & 20.36/57.21  & {\bf21.97/60.82} \\ %[-0.5ex]
\hline
\end{tabular}
\caption{Comparison of the denoising performance in terms of PSNR and 
SSIM at various noise levels using the \textbf{box} kernel
% . The parameters used are $S = 10$ and $K=3$. 
with parameters $S = 10$ and $K=3$.
The images used here are: 
\emph{M1: Man}, \emph{B1: Barbara}, \emph{H1: House}, \emph{B2: Bridge}, 
and \emph{C1: Cameraman}. The images  \emph{M1} and \emph{B1} are of 
size $512 \times 512$,  while the rest of the images are of size $256 
\times 256$.}
\label{psnrBox}
}
\end{table*}

%\newpage
\section{Conclusion}
\label{sec:conclusion}

We proposed a separable formulation of non-local means and a fast algorithm for the same. The algorithm admits a simple implementation for both box and Gaussian kernels. The overall complexity of the algorithm is smaller than that of existing fast implementations of NLM. In fact, the actual run-time of the Matlab implementation of the algorithm is competitive with the run-time of existing implementations, particularly when the patch size and the search window are large. The speedup over NLM was observed to be at least $300$ times. We also demonstrated that the denoising performance of the proposed algorithm is consistently better than NLM and some of its variants in terms of PSNR/SSIM and the visual quality. It is clear that the SNLM has a straightforward extension to video and volume data, where the acceleration would be even more significant. As a final remark, we note that the rapidly computed SNLM can be used to seed various sophisticated methods for image denoising that require some kind of initialization (such as dictionary-based denoising schemes \cite{KSVD}). It can also be used to seed the Wiener filter used in second stage of the BM3D algorithm \cite{BM3D} that yields state-of-the-art results.

\section{Appendix}

In this section, we describe the main steps in the derivation of \eqref{div}. The steps are similar to that in \cite{VK2009}. In particular, by applying the quotient rule for derivatives to \eqref{1dNLM}, we have
\begin{align}
\label{derivfrac} 
\frac{\partial \hat{f}(i)}{\partial f(i)} &=  \frac{1}{W_i} \Big( w_{i i} +  \sum_{j=i-S}^{i+S}  f(j)  \frac{\partial w_{ij}}{\partial f(i)}    - \hat{f}(i) \sum_{j=i-S}^{i+S}  \frac{\partial w_{ij}}{\partial f(i)} \Big) \nonumber \\
&= \frac{1}{W_i} \Big( w_{i i} +  \sum_{j=i-S}^{i+S} ( f(j) - \hat{f}(i) )  \frac{\partial w_{ij}}{\partial f(i)} \Big),
\end{align}
where $W_i = \sum_{j=i-S}^{i+S} w_{i j}$. Using the chain rule for derivatives, we note that for  $|j-i| > K$,
\begin{equation*}
  \frac{\partial w_{ij}}{\partial f(i)} =    \frac{2}{ h^2}   w_{ij} g_{\beta}(0) \big( {f(j) - f(i)} \big).
\end{equation*}
On the other hand, for $|j-i| \leq K$,
 \begin{equation*}
 \frac{\partial w_{ij}}{\partial f(i)} = \frac{2 w_{ij}}{ h^2} \big(g_{\beta}(0) ( {f(j) - f(i)})  + g_{\beta}(i-j) ({f(2i-j) - f(i)} ) \big).
\end{equation*}
On plugging the above in \eqref{derivfrac}, and noting that $w_{ii}=1$ and $g_{\beta}(0) =1$, we get \eqref{div}.

\section{Acknowledgements}

The authors would like to thank the anonymous reviewers for their valuable comments and suggestions. The authors are also grateful to Tolga Tasdizen for sharing the Matlab implementation of principal neighborhood dictionary with the research community.

\bibliographystyle{IEEEbib}

\begin{thebibliography}{9}


%%%%%%%%%%%%%%%%%%%%%%%%%%%%%
\bibitem{Buades2005} A. Buades, B. Coll, and J.-M. Morel, ``A non-local algorithm for image denoising," \textit{Proc. IEEE Conference on Computer Vision and Pattern Recognition}, {\bf2}, pp. 60-65 (2005).

\bibitem{Kervrann2006} C. Kervrann and J. Boulanger, ``Optimal spatial adaptation for patch-based image denoising,'' \textit{IEEE Transactions on Image Processing}, {\bf15}(10), pp. 2866-2878 (2006).

\bibitem{KSVD} M. Elad and M. Aharon, ``Image denoising via sparse and redundant representations over learned dictionaries,''  \textit{IEEE Transactions on Image Processing}, {\bf15}(12), pp. 3736-3745 (2006).

\bibitem{BM3D} K. Dabov, A. Foi, V. Katkovnik, and K. Egiazarian, ``Image denoising by sparse 3-D transform-domain collaborative filtering,'' \textit{IEEE Transactions on Image Processing}, {\bf16}, pp. 2080-2095 (2007).

\bibitem{Giloba2008} G. Gilboa and S. Osher, ``Nonlocal operators with applications to image processing,'' \textit{Multiscale Modeling and Simulation}, {\bf7}(3), 1005-1028 (2008).

\bibitem{Chatterjee2012} P. Chatterjee and P. Milanfar, ``Patch-based near-optimal image denoising,'' \textit{IEEE Transactions on Image Processing}, {\bf21}(4), pp. 1635-1649 (2012).

\bibitem{BCM2010} A. Buades, B. Coll, and J. M. Morel, ``Image denoising methods. A new nonlocal principle,'' \textit{SIAM Review}, {\bf52}, pp. 113-147 (2010).

\bibitem{Darbon2008} J. Darbon,  A. Cunha,  T. F. Chan, S. Osher,  and G. J. Jensen, ``Fast nonlocal filtering applied to electron cryomicroscopy,''  \textit{Proc. IEEE International Symposium on Biomedical Imaging}, pp. 1331-1334 (2008).

\bibitem{Tasdizen2008} T. Tasdizen, ``Principal components for non-local means image denoising,'' \textit{Proc. IEEE International Conference on Image Processing}, pp. 1728-1731 (2008).

\bibitem{Tasdizen2009} T. Tasdizen, ``Principal Neighborhood Dictionaries for Nonlocal Means Image Denoising,'' \textit{IEEE Transactions on Image Processing}, {\bf18}(12), pp. 2649-2660 (2009).

\bibitem{WTNN2013} Y. Wu, B. Tracey, P. Natarajan, and J. Noonan, ``Probabilistic non-local means,'' \textit{IEEE Signal Processsing Letters}, {\bf20}(8), pp. 763-766 (2013).

\bibitem{Yaroslavsky1985} L. P. Yaroslavsky, \textit{Digital Picture Processing}. Secaucus, NJ: Springer-Verlag (1985).

\bibitem{Smith1997} S. M. Smith and J. M. Brady, ``SUSAN - A new approach to low level image processing,'' \textit{International Journal of Computer Vision}, {\bf23}(1), pp. 45-78 (1997).

\bibitem{Tomasi1998} C. Tomasi, R. Manduchi, ``Bilateral filtering for gray and color images,'' \textit{Proc. IEEE International Conference on Computer Vision}, pp. 839-846 (1998).


\bibitem{VJ2001} P. Viola and M. Jones, ``Rapid object detection using a boosted cascade of simple features," \textit{Proc. IEEE Conference on Computer Vision and Pattern Recognition}, {\bf1}, pp. 511-518 (2001).

\bibitem{Wang2006} J. Wang, Y. Guo, Y. Ying, Y. Liu, and Q. Peng, ``Fast non-local algorithm for image denoising,'' \textit{Proc. IEEE International Conference on Image Processing}, pp. 1429-1432 (2006).

\bibitem{KUD2009} V. Karnati, M. Uliyar, and S. Dey, ``Fast non-local algorithm for image denoising,'' \textit{Proc. IEEE International Conference on Image Processing}, pp. 3873-3876 (2009).

\bibitem{Condat2010} L. Condat, ``A simple trick to speed up and improve the non-local means, '' \textit{Research Report} HAL-00512801, (2010).

\bibitem{Sapiro2005} M. Mahmoudi and G. Sapiro, ``Fast image and video denoising via nonlocal means of similar neighborhoods,'' \textit{IEEE Signal Processing Letters}, {\bf12}(12), pp. 839-842 (2005).

\bibitem{Coupe2008} P. Coupe, S. Prima, P. Hellier, C. Kervrann, and C. Barillot, ``An optimized blockwise nonlocal means denoising filter for 3-D magnetic
resonance images,'' \textit{IEEE Transactions on Medical Imaging}, {\bf27}(4), pp. 425-441 (2008).

\bibitem{Brox2008} T. Brox, O. Kleinschmidt, and D. Cremers, ``Efficient non-local means for denoising of textural patterns,'' \textit{IEEE Transactions on Image Processing}, {\bf17}(7), pp. 1083-1092 (2008).

\bibitem{Dauwe2008} A. Dauwe, B. Goossens, H. Luong, and W. Philips, ``A fast non-local image denoising algorithm,'' \textit{Proc. SPIE Electronic Imaging}, {\bf 6812}, pp. 1331-1334 (2008).

\bibitem{earlytermination2010} R. Vignesh, O. T. Byung, and C. C. Kuo, ``Fast non-local means (NLM) computation with probabilistic early termination,'' \textit{IEEE Signal Processing Letters}, {\bf17}(3), pp. 277-280 (2010).

\bibitem{Orchard2008} J. Orchard, M.  Ebrahimi, and A. Wong, ``Efficient nonlocal-means denoising using the SVD,'' \textit{Proc. IEEE International Conference on Image Processing}, pp. 1732-1735 (2008).

\bibitem{Adams2009} A. Adams, N. Gelfand, J. Dolson, and M. Levoy, ``Gaussian KD-trees for fast high-dimensional filtering,'' \textit{Proc. ACM Siggraph}, (2009).

\bibitem{Adams2010} A. Adams, J. Baek, and A. Davis, ``Fast high-dimensional filtering using the permutohedral lattice,'' \textit{Proc. EuroGraphics}, (2010).

\bibitem{Deledalle2012} C. Deledalle, V. Duval, and J. Salmon, ``Non-local methods with shape-adaptive patches (NLM-SAP),'' \textit{Journal of Mathematical Imaging and Vision},  {\bf43}(2), pp. 103-120 (2012).

\bibitem{BC2014} ``Novel Speed-Up Strategies for Non-Local Means Denoising With Patch and Edge Patch Based Dictionaries,'' \textit{IEEE Transactions on Image Processing}, {\bf23}(1), pp. 356-364 (2014).

\bibitem{N1981} P. M. Narendra, ``A separable median filter for image noise smoothing,'' \textit{IEEE Transactions on Pattern Analysis and Machine Intelligence}, {\bf3}, pp. 20-29 (1981).

\bibitem{Pham2005} T. Q. Pham and L. J. Van Vliet, ``Separable bilateral filtering for fast video preprocessing,'' \textit{Proc. IEEE International Conference on Multimedia and Expo} (2005).

\bibitem{KLCLKK2011} Y. S. Kim, H. Lim, O. Choi, K. Lee, J. D. K. Kim, and C. Kim, ``Separable bilateral non-local means,'' \textit{Proc. IEEE International Conference on Image Processing}, pp. 1513-1516 (2011).


%\bibitem{Chaudhury2013} K. N. Chaudhury, ``Acceleration of the shiftable algorithm for bilateral filtering and nonlocal means,'' \textit{IEEE Transactions on Image Processing}, vol. 22, no. 4, pp. 1291-1300, 2013.

%\bibitem{Chaudhury2011a} K. N. Chaudhury, ``Constant-time filtering using shiftable kernels,'' \textit{IEEE Signal Processing Letters}, vol. 18, no. 11, pp. 651-654, 2011.

\bibitem{S1981} C. Stein, ``Estimation of the mean of a multivariate normal distribution,'' {\em Annals of Statistics,} {\bf 9}, pp. 1135-1151 (1981).

\bibitem{BL2007} T. Blu and F. Luisier, ``The SURE-LET approach to image denoising'', {\em  IEEE Transactions on Image Processing},  {\bf 16}(11),  pp.2778-2786 (2007).

\bibitem{D1992} R. Deriche, ``Recursively implementing the Gaussian and its derivatives'', \textit{Proc. IEEE International Conference on Image Processing},  pp. 263-267 (1992). 

\bibitem{YV1995} I. T. Young and L. J. Van Vliet, ``Recursive implementation of the Gaussian filter'',  \textit{Signal Processing},   {\bf44},  pp. 139-151 (1995). 

\bibitem{VK2009} D. Van De Ville and M. Kocher, ``Sure-based non-local means,'' {\em IEEE Signal Processing  Letters}, {\bf16}(11), pp. 973-976 (2009).

\bibitem{Chaudhury2011} K. N. Chaudhury, D. Sage, and M. Unser, ``Fast bilateral filtering using trigonometric range kernels,'' \textit{IEEE Transactions on Image Processing}, {\bf20}(12), pp. 3376-3382 (2011).

\bibitem{BM3Dimages} \url{http://www.cs.tut.fi/~foi/GCF-BM3D}.

\bibitem{SIPI} \url{http://sipi.usc.edu/database}.

\bibitem{IPOL} \url{http://www.ipol.im}.

\bibitem{SSIM2004} Z. Wang, A. C. Bovik, H. R. Sheikh, and E. P. Simoncelli, ``Image quality assessment: From error visibility to structural similarity,'' \textit{IEEE Transactions on Image Processing}, {\bf13}(4), pp. 600-612 (2004).
\end{thebibliography}

\end{document}